\documentclass[journal]{IEEEtai}

\usepackage[colorlinks,urlcolor=blue,linkcolor=blue,citecolor=blue]{hyperref}

\usepackage{color,array}
\usepackage{amsmath,amsfonts}
\usepackage{algorithmic}
\usepackage{array}
\usepackage[caption=false,font=normalsize,labelfont=sf,textfont=sf]{subfig}
\usepackage{textcomp}
\usepackage{stfloats}
\usepackage{url}
\usepackage{verbatim}
\usepackage{graphicx}
\usepackage{bbm}
\usepackage{times}
\usepackage{soul}
\usepackage{bm}
\usepackage{url}
\usepackage{multirow}
\usepackage{multicol}

\usepackage[small]{caption}
\usepackage{graphicx}
\usepackage{amsmath}
\usepackage{amsthm}
\usepackage{booktabs}
\usepackage{algorithm}
\usepackage{algorithmic}
\usepackage{stfloats}
\usepackage{color}
\usepackage{amsmath}
\usepackage{amssymb}

\usepackage{adjustbox}
\usepackage{graphicx}
\usepackage{balance}

\begin{document}

\title{Inaccurate Label Distribution Learning}

 \author{Zhiqiang Kou,~Yuheng~Jia,~\IEEEmembership{Member,~IEEE},~Jing~Wang\and,~Xin Geng,~\IEEEmembership{Senior Member,~IEEE,}\\

  \thanks{The authors are with the School of Computer Science and Engineering, Southeast University, Nanjing 211189, China, and also with the Key Laboratory of Computer Network and Information Integration (Southeast University), Ministry of Education, Nanjing 211189, China.}
\thanks{Corresponding author: Xin Geng and Yuheng Jia.}}

\markboth{Journal of IEEE Transactions on Artificial Intelligence, Vol. 00, No. 0, Month 2020}
{First A. Author \MakeLowercase{\textit{et al.}}: Bare Demo of IEEEtai.cls for IEEE Journals of IEEE Transactions on Artificial Intelligence}

\maketitle

\begin{abstract}
Label distribution learning (LDL) trains a model to predict the relevance of a set of labels (called label distribution (LD)) to an instance. The previous  LDL methods all assumed the LDs of the training instances are accurate.  However, annotating highly accurate  LDs for training instances  is time-consuming and very expensive, and in reality the collected LD is usually inaccurate and disturbed by annotating errors.  For the first time, this paper investigates the problem of inaccurate LDL, i.e., developing an LDL model with noisy LDs.  We assume that the noisy LD matrix is a linear combination of an ideal LD matrix and a sparse noise matrix. Consequently, the problem of inaccurate LDL becomes an inverse problem, where the objective is to recover the ideal LD and noise matrices from the noisy LDs. We hypothesize that the ideal LD matrix is low-rank due to the correlation of labels and utilize the local geometric structure of instances captured by a graph to assist in recovering the ideal LD. This is based on the premise that similar instances are likely to share the same LD. The proposed model is finally formulated as a graph-regularized low-rank and sparse decomposition problem and numerically solved by the alternating direction method of multipliers. Furthermore, a specialized objective function is utilized to induce a LD predictive model in LDL, taking into account the recovered label distributions. 
Extensive experiments conducted on multiple datasets from various real-world tasks effectively demonstrate the efficacy of the proposed approach.
\end{abstract}

\begin{IEEEImpStatement}
LDL has gained popularity among researchers for addressing label ambiguity problems and yielding promising results. It provides precise supervision information for finer-grained predictions. However, accurate labeling of training instances is time-consuming and expensive, leading to inaccuracies and noise in real-world scenarios. To address this challenge, this paper introduces a novel method based on graph-regularized low-rank and sparse decomposition. Our method enhances model robustness against label distribution noise, ensuring reliable performance in challenging conditions. It has the potential to support various LDL methods, including facial expression recognition, facial age estimation, and other intelligent detection and recognition scenarios.
\end{IEEEImpStatement}


\section{Introduction}
\IEEEPARstart{L}{abel}  distribution learning (LDL) is an emerging topic in machine learning. Different from the traditional single-label learning and multi-label learning, which use binary value to specify whether an instance is related to a certain label, LDL solves the problem of to what degree a label can describe an instance. This powerful learning paradigm is good at handling label ambiguity and has many real-world applications, like  music classification \cite{icassp/BuissonAB22},  breast tumor cellularity assessment \cite{miccai/LiLLWWL22}, and facial age estimation \cite{wen2020adaptive}.

\begin{figure}[t]
\centering
\includegraphics[width=0.5\textwidth]{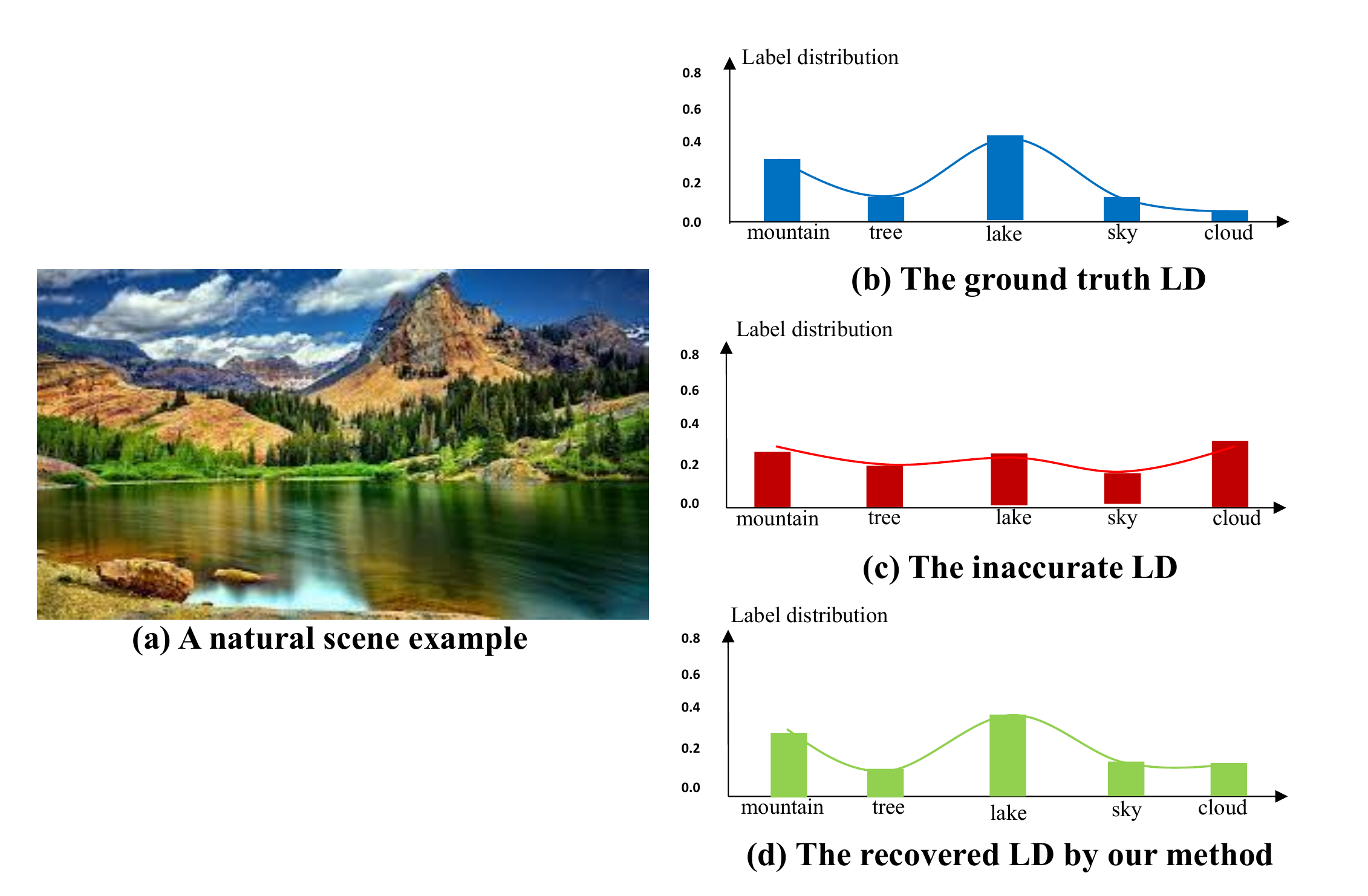} 
\caption{Illustration of the inaccurate label distribution learning  problem. (a) denotes is a natural scene image, (b) and (c) denote the  correct label distribution and  inaccurate label distribution, and (d) indicates   the label distribution recovered by the proposed method from the inaccurate label distribution.}
\label{fig_amb_distribution}
\end{figure}
LDL was first proposed by Geng \cite{geng2016label}. In LDL, the relative importance of each label to an instance is called the description degree, which is captured by a label distribution (LD).  Fig. 1(a) shows a multi-label scene image, where “lake” has higher importance than “cloud”, and at the same time both of them are positive labels, so it makes sense to know the description  degree  of each label, which forms an LD as shown in Fig. 1(b). Similar to other machine learning paradigms, in the training phase of LDL, a training set with many instances  and the annotated LDs are given to train an LDL model. In the test phase, the learned LDL model is used to predict the LD for an unseen  sample. 

To solve the LDL problem, different models have been proposed. For example, Jia et al. \cite{jia2019label}   used label correlations on local samples and proposed two new LDL algorithms, called GD-LDLSCL and Adam-LDL-SCL, respectively. In order to solve the objective mismatch and improve the classification performance of LDL,  
 Wang and Geng \cite{wang2021label}  proposed  the label distribution learning machine. To reduce the high computational overhead of LDL, Tan et al \cite{tan2020multilabel} developed an LDL algorithm based on stream learning with multiple output regression, called MDLRML. To avoid the problem that LDL treats data differently in the training stage and testing stage,  Wang et al. \cite{wang2021re}  proposed the re-weighting large margin label distribution learning.  Considering the label ranking relationships, Jia et al. \cite{jia2021label} introduced  a ranking loss function to the traditional LDL models. 

\textbf{Motivation}: Although those LDL methods have achieved great success in many applications, all of them assumed that the LDs of the training instances are accurate, however,  precisely assigning an accurate LD to an instance is extremely time-consuming and expensive. Therefore,  in reality, the LDs collected in the training set are usually inaccurate with many noises, and inaccurate LDs become a common phenomenon in LDL.  For example, Fig. 1(b)  shows the ground truth LD  of an instance,  and Fig. 1(c) denotes  the inaccurate LD, which   puts higher description degree to the label “colud" and lower description degree to the label “lake".  It is very important to investigate how to construct a reliably LDL model with inaccurate LDs, which unfortunately has been overlooked by the previous researches. 

In this paper, we study the problem of inaccurate label distribution learning (ILDL), i.e.,  design an LDL model with a training set annotated by inaccurate  LD,  for the first time. Specifically, we assume that the inaccurate LD is the linear combination of an ideal LD and a sparse noise. Then, the ILDL  problem can be treated as an inverse problem to separate the ideal LD and the noise label  from the inaccurate observations. To this end, we collect the LDs of all the instances to construct an LD matrix and  assume that the ideal LD matrix is low-rank, since the labels are usually correlated to each other in multi-label learning \cite{xu2017incomplete}, and the noise label  matrix is sparse due to the fact that the coarse labeling usually generates only a small fraction of noise.  Moreover,  if  two instances  are similar enough,  their LDs should also be similar to each other.  Motivated by this observation, we use the local similarity structure of the instrances  to assist the recovery of the ideal LD. Finally, we formalize the proposed model as a low-rank and sparse decomposition  problem  with a graph regularization  (LSag)   and solve it using the alternating direction method of multipliers (ADMM) \cite{boyd2011distributed}. The recovered LD are taken into consideration when inducing a LD predictive model for LDL, achieved through the utilization of a specialized objective function. Extensive experiments  validate the advantage of our approach over the state-of-the-art approaches. 

We organize the rest of the paper as follows. First, related
works on LDL are briefly discussed. Second, technical details
of the proposed approach are introduced. Third, experimental
results of comparative studies are reported. Finally, we conclude this paper.

\section{Related Work}

\subsection{Lable Distribution Learning}
 As a new learning paradigm, LDL can better describe the labeling degree of an instance than the traditional multi-label learning.  Accordingly, LDL has attracted a lot of attention.  In this section, we briefly review the researches in LDL. 

The develop of LDL is inspired by solving various  real-world applications. For example,  in the early years, LDL shined in facial age recognition task \cite{geng2013facial}.  After that, Geng 
 \cite{geng2014head} proposed an LDL-based head pose estimation algorithm, which makes full use of the multi-label distribution information. Zhou et al. \cite{zhou2015emotion} found that all facial expressions in nature cannot be defined by a binary label, and accordingly, they   developed a facial emotion recognition algorithm based on LDL. In addition, the idea of  LDL has been applied to the prediction  of multi-component compositions of Martian craters \cite{Morrison2018PredictingMM}, age estimation  of the speaker \cite{icassp/SiWPX22},  indoor crowd counting  \cite{ling2019indoor}, and infant age estimation  \cite{hu2019deep}.

Apart from the real-world applications, many  researches focus on developing an effective LDL model  for general purposes. We roughly divide the  existing  LDL algorithms  into three categories. The first category converts the LDL problem into a single-label learning problem, i.e.  transforming  the training samples into a set of weighted single-label samples. The representative algorithms are PT-SVM and PT-Bayes \cite{geng2016label}, which use the SVM algorithm and the Bayes classifier to solve the transformed weighted single-label learning problem.  The second category is  algorithm adaption, which extends the traditional machine learning algorithms  to deal with the LDL problem. For example, the K-nearest neighbors (KNN) classifier finds the top $k$ neighbors of an instance and uses the average labels of the top $k$ neighbors as the prediction of the LD of that instance. Backpropagation (BP) neural networks can directly  minimizes the descriptive degree of the final prediction through the BP algorithm.  The last category is specialized algorithms, such as IIS-LDL and BFGS-LDL \cite{geng2013facial}. They formulated LDL as a  regression problem and  used an improved iterative scaling algorithm and a quasi-Newton method to solve the final regression problem, respectively.

As the LDs are usually annotated by different persons with diverse  levels of experience, assigning  a precise description degree to all instances is very challenging, and inaccurate LD is a common phenomenon in LDL. However, the previous researches all assumed  the LD of the training set is accurate, which cannot handle  the inaccurate LDL problem. 
This paper will investigate  the inaccurate LDL problem for the first time. 

\section{The Proposed Method }
\textbf{Notations:}  Let $\mathbf{X}=\left[\mathbf{x}_{1}, \mathbf{x}_{2}, \ldots, \mathbf{x}_{n}\right] \in \mathbbm{R}^{n \times d}$ denote the feature matrix, and $Y=\left\{y_{1}, y_{2}, \ldots, y_{m}\right\}$ be the label space,  where $n$, $m$, and $d$  denote the number of instances,  the number of the labels and the dimension of features. The training set of the LDL problem is represented as: $\mathbbm{T}=\left\{\left(\mathbf{x}_{1}, \mathbf{d}_{1}\right), \quad\left(\mathbf{x}_{2}, \mathbf{d}_{2}\right), \ldots,\left(\mathbf{x}_{n}, \mathbf{d}_{n}\right)\right\}$, 
where $\mathbf{d}_{i}=\left[d_{\mathbf{x}_{i}}^{y_{1}}, d_{\mathbf{x}_{i}}^{y_{2}}, \ldots, d_{\mathbf{x}_{i}}^{y_{m}}\right]$ is the label distribution vector to the $i$th sample $\mathbf{x}_{i}$.  $d_{\mathbf{x}_{i}}^{y}$ indicates the importance degree of label $y$ to $\mathbf{x}_{i}$, which satisfies  $d_{\mathbf{x}_i}^{y} \in[0,1]$ and $\sum_{y} d_{x}^{y}=1$. The LD matrix of all the instances  is denoted as $\mathbf{D}=\left[\mathbf{d}_{1}, \mathbf{d}_{2}, \ldots, \mathbf{d}_{n}\right] \in \mathbbm{R}^{n \times m}$.  LDL aims to   learn a mapping function from $\mathbbm{T}$, which can predict the LD for unseen instances.

The traditional LDL approaches  all assume  that LD matrix $\mathbf{D}$ is accurate, but considering the fact that precisely annotating the LD for an instance is very costly, in reality,  the  collected LD matrix is usually not accurate, which is polluted by labeling noise.  Directly training an LDL model with noisy LD will certainly result in unsatisfactory performance.  To this end, this paper  investigates the problem of inaccurate LDL, which can construct a reliable LDL model from the noisy LD.
\subsection{Low-rank and Sparse Decomposition of the Noisy Label Distribution}
To achieve inaccurate LDL, we assume that the  observed noisy   LD matrix is the linear combination of an ideal LD matrix and labeling error matrix, i.e.,
\begin{equation}
\mathrm{\mathbf{D}=\mathbf{{\widetilde{D}}}+\mathbf{E},}
\end{equation}
where $\mathbf{\widetilde{D}}\in \mathbbm{R}^{n \times m}$ denotes the to be recovered  ideal LD matrix, $\mathbf{E} \in \mathbbm{R}^{n \times m}$  represents the error term in the LD.  Accordingly, the inaccurate LDL problem becomes an inverse problem, i.e., recovering the ideal LD matrix $\mathbf{{\widetilde{D}}}$ and the error matrix $\mathbf{E}$ from the inaccurate LD matrix $\mathbf{D}$. 

To solve the inverse  problem, we need to leverage the characteristics of the ideal LD matrix and error matrix. In LDL, each instance has multiple valid labels, and the label correlations exist in most multiple label learning problems. Due to label correlations, the ideal LD matrix is supposed to be low-rank. Note that the low-rankness of the LD matrix has been verified in \cite{xu2017incomplete}. Besides, although  the given LD is not accurate, it is usually annotated by different persons with some training on annotation, we assume that only minority proportion of the LDs is inaccurate, and accordingly, the error matrix is sparse. Based on the above assumptions, the proposed ILDL problem is preliminarily formulated as
\begin{equation}
\begin{aligned}
& \min_{\mathbf{\widetilde{D}}, \mathbf{E} } \operatorname{rank}(\mathbf{\widetilde{D}})+\alpha\operatorname{card}(\mathbf{E}) \\ 
& \text { s.t. } \mathbf{D}=\mathbf{\widetilde{D}}+\mathbf{E},
\end{aligned}
\label{firstloss}
\end{equation}
where $\operatorname{rank}(\mathbf{\widetilde{D}})$ denotes the rank of the ideal LD matrix,  $\operatorname{card(\mathbf{E})}$ records the number of non-zero elements in $\mathbf{E}$, and $\alpha$ is the trade-off parameter. By solving Eq. (\ref{firstloss}), the noisy LD matrix $\mathbf{D}$ will be decomposed to a low-rank ideal LD matrix $\mathbf{\widetilde{D}}$ and a sparse error term $\mathbf{E}$. 
\subsection{Exploiting Instances Correlations by Adaptive Graph Learning}
The correlations among the instances are  also important for recovering the ideal LD, i.e. if two instances are close in feature space,  their ideal LDs should also be similar to each other.  In order to capture the similarity relationships  of the instances, we construct an adaptive graph $\mathbf{A}\in \mathbbm{R}^{n \times n}$ as:
\begin{equation}
\begin{aligned}
&\min _{{a_i}} \sum_{j=1}^n\left(\frac{1}{2}\left\|\mathbf{x}_i-\mathbf{x}_j\right\|_2^2 a_{i j}+\gamma a_{i j}^2\right)\\
&\text { s.t. } {a_i}^T \mathbf{1}_n=\mathbf{1}_n, \forall i.j, 0 \leq a_{i j} \leq 1,
\end{aligned}
\label{graph}
\end{equation}
where  $a_{i j}$ is the $(i,j)$-th element of $\mathbf{A}$, which represents the similarity between  $\mathbf{x}_{i}$ and $\mathbf{x}_{j}$, $\mathbf{1}_n\in \mathbbm{R}^{1 \times n}$ is an all ones vector with size $n$, and $\gamma>0$ is a trade-off parameter. The first term in Eq. (\ref{graph}) ensures that ${a}_{i j}$  is larger when $\mathbf{{x}}_{i}$ and $\mathbf{{x}}_{j}$ are similar  to each other.  The second term of Eq. (\ref{graph}) avoids the trivial solution, i.e., $\mathbf{A}$ becomes an identity matrix. The constraints of Eq. (\ref{graph}) guarantee that the similarities among instances are non-negative and the similarity matrix is normalized. After solving Eq. (\ref{graph}), we use the similarity relationship $\mathbf{A}$ of samples to guide the ideal LD recovery, i.e.,

\begin{equation}
    \sum_{i,j} \min  a_{i j}\left\|\widetilde{\mathbf{d}}_{i}\!-\!\widetilde{\mathbf{d}}_{j}\right\|^{2} \!=\!\operatorname{Tr}\left(\widetilde{\mathbf{D}} \mathbf{L} \widetilde{\mathbf{D}}^{\mathrm{T}}\right),
    \label{traceDLD}
\end{equation}
 where $\operatorname{Tr}(\cdot)$  is the trace of a matrix. $\mathbf{L}=\mathbf{\hat{A}}+\left(\mathbf{A}+\mathbf{A}^{\top}\right)/2\in \mathbbm{R}^{n\times n}$ is the graph Laplacian matrix, $\mathbf{\hat{A}}$ is a diagonal matrix with the $(i,i)$-th element  $\mathbf{\hat{A}}_{i i}=\sum_{j=1}^n\left[\left(a_{i j}+a_{j i}\right) / 2\right]$. By minimizing Eq. (\ref{traceDLD}), two instances with similar feature representations will tend to own the similar LDs.

\subsection{Model Formulation}

Combining  the above priors, our model is formulated as:
\begin{equation}
\begin{aligned}
& \min_{\mathbf{\widetilde{D}}, \mathbf{E} } \operatorname{rank}(\mathbf{\widetilde{D}})+\alpha\operatorname{card}(\mathbf{E})+\beta\operatorname{Tr}\left(\widetilde{\mathbf{D}} \mathbf{L} \widetilde{\mathbf{D}}^{\mathrm{T}}\right)\\ 
& \text { s.t. } \mathbf{D}=\mathbf{\widetilde{D}}+\mathbf{E},
\end{aligned}
\label{dao2loss}
\end{equation}
where $\beta$ is the trade-off parameter. As the rank function $\operatorname{rank}(\cdot$) and the  card function $\operatorname{card}(\cdot$) are both non-convex and discrete, Eq. (\ref{dao2loss}) is difficult to solve. Therefore, we relax those two terms by the associated  convex surrogates, i.e. nuclear norm for $\operatorname{rank}(\cdot)$ and $\ell_{1}$ norm for $\operatorname{card}(\cdot)$, and our model finally becomes
\begin{equation}
\begin{aligned}
& \min_{\mathbf{\widetilde{D}}, \mathbf{E} } \alpha\|\mathbf{E}\|_{1} +\|\mathbf{\mathbf{\widetilde{D}}}\|_{*}+\beta\operatorname{Tr}\left(\widetilde{\mathbf{D}} \mathbf{L} \widetilde{\mathbf{D}}^{\mathrm{T}}\right)\\ 
& \text { s.t. } \mathbf{D}=\mathbf{\widetilde{D}}+\mathbf{E}.
\end{aligned}
\label{finnalloss}
\end{equation}
By solving  Eq. (\ref{finnalloss}), we can recover  an ideal LD and a sparse error matrix from the noise LD. Then any LDL algorithms can be applied on $\widetilde{\mathbf{D}}$ to learn a reliable label distribution prediction model.

\subsection{Making Prediction }
After solving Eqs.  (\ref{graph}) and (\ref{finnalloss}), a clean LD is learned, since $\mathbf{d}_i$ is a real-valued quantity, multi-output support vector regression (MSVR) \cite{chung2014general}, \cite{perez2002multi} is utilized to address this scenario. In this approach, a kernel regression model is employed to parameterize the label distribution predictor:

\begin{equation}
\begin{aligned}
& \min _{\mathbf{(\Theta, b)}} \frac{1}{2}\|\mathbf{\Theta}\|_F^2+\kappa \ell(\mathbf{(\Theta, b)}) \\
& \text { s.t } \forall i, \widetilde{d}_i^j\left(\boldsymbol{\theta}_j^{\top} \varphi\left(\boldsymbol{x}_i\right)+b_j\right) \geqslant 0,
\end{aligned}
\label{fenleimodel}
\end{equation}
where $\boldsymbol{\Theta}=\left[\boldsymbol{\theta}_1, \boldsymbol{\theta}_2, \ldots, \boldsymbol{\theta}_q\right]$ and $\boldsymbol{b}=\left[b_1, b_2, \ldots, b_q\right]^{\top}$ signify the weight matrix and the bias vector of the regression model, respectively. As indicated in Eq. (\ref{fenleimodel}), the first term is responsible for regulating the complexity of the resulting model. The second term represents the hinge loss, and its specific definition is as follows: $\ell(\mathbf{(\Theta, b)}=\max (0, u_i-\varepsilon)$, here $u_i=\left\|\boldsymbol{e}_i\right\|=\sqrt{\boldsymbol{e}_i^{\top} \boldsymbol{e}_i}$ with $\boldsymbol{e}_i=\boldsymbol{\widetilde{d}}_i-\boldsymbol{\Theta}^{\top} \varphi\left(\boldsymbol{x}_i\right)-\mathbf{b}$. The hinge loss generates an insensitive zone around the estimation, determined by $\epsilon$. In other words, any loss of $u_i$ smaller than $\epsilon$ will be disregarded. The constraint is employed to maintain consistency between the signs of the prediction and the ideal LD matrix $\widetilde{d}_i^j$. In order to facilitate the optimization of the objective function, we relax the constraint to: $\forall i, \widetilde{d}_i^j\left(\boldsymbol{\theta}_j^{\top} \varphi\left(\boldsymbol{x}_i\right)+b_j\right) \geqslant 0$ = $-\sum_{i=1}^n \sum_{j=1}^c \widetilde{d}_{\boldsymbol{x}_i}^j \boldsymbol{\theta}_j^{\top} \boldsymbol{\phi}_i$ = $-\operatorname{tr}\left(\hat{\mathbf{L}}^{\top} \mathbf{\Theta \Phi}\right)$, where $\boldsymbol{\Phi}=\left[\phi_1, \phi_2, \ldots, \phi_n\right]$.

\subsection{Numerical Solution of Eq. (\ref{finnalloss})}
We use  ADMM to solve problem (\ref{finnalloss}),  which is good at handling the equality constraints. First,  we introduce an intermediate variable $\mathbf{Z}\in \mathbbm{R}^{n \times m}$,  and rewrite Eq. (\ref{finnalloss}) as :
\begin{equation}
\begin{aligned}
& \min_{\mathbf{\widetilde{D}}, \mathbf{E} } \alpha\|\mathbf{E}\|_{1} +\|\mathbf{\mathbf{Z}}\|_{*}+\beta\operatorname{Tr}\left(\widetilde{\mathbf{D}} \mathbf{L} \widetilde{\mathbf{D}}^{\mathrm{T}}\right)\\ 
& \text { s.t. } \mathbf{D}=\mathbf{\widetilde{D}}+\mathbf{E},   \mathbf{\widetilde{D}}=\mathbf{Z} .
\end{aligned}
\label{VAZ}
\end{equation}
Then,  the  augmented Lagrangian form of Eq. (\ref{VAZ}) is:
\begin{equation}\footnotesize
\begin{aligned}
 &\mathcal{L}_{(\mathbf{\widetilde{D}, E,  Z, \Gamma_{1}, \Gamma_{2}})}
 \!=\!\beta\operatorname{Tr}\left(\mathbf{\widetilde{D} L \widetilde{D}^{T}}\right)\!+\!\alpha\|\mathbf{E}\|_{1}\!+\!\|\mathbf{Z}\|_{*}\!+\!\langle\mathbf{\Gamma}_{2}, \mathbf{\widetilde{D}-Z}\rangle\\
& +\langle\mathbf{\Gamma}_{1}, \mathbf{\widetilde{D}+E-D}\rangle +\frac{\mu}{2}\left(\|\mathbf{\widetilde{D}+E-D\|}_{F}^2+\mathbf{\|\widetilde{D}-Z\|}_{F}^{2}\right) 
 ,\\
\end{aligned}
\label{alage}\footnotesize
\end{equation}
where $\mathbf{\Gamma}_{1}\!\in\! \mathbbm{R}^{n \times m}$, $\mathbf{\Gamma}_{2}\!\in\! \mathbbm{R}^{n \times m}$ denote the Lagrangian multipliers, $\mu$ is a positive penalty parameter, $\|\cdot\|_{F}^2$ is Frobenius norm, $\langle \cdot \rangle$ denotes the inner product of two vectors. Eq. (\ref{alage}) can be solved by alternately solving the following sub-problems:
\begin{enumerate}
    \item $\mathbf{Z}$-Subproblem is formulated as:
    \begin{equation}
    \begin{aligned}
         \min _\mathbf{Z}\|\mathbf{Z}\|_{*}+\langle\mathbf{\Gamma}_{2}, \mathbf{\widetilde{D}-Z}\rangle 
         +\frac{\mu}{2}\|\mathbf{\widetilde{D}-Z}\|_{F}^{2}.
    \end{aligned}
    \label{Zsub}
\end{equation}

Eq. (\ref{Zsub}) is a nuclear norm minimization problem, with a closed-form solution, i.e, \cite{cai2010singular}
\begin{equation}
\begin{aligned}
\mathbf{Z}^{k+1}=\mathcal{J}_{1 / \mu}\left(\mathbf{\widetilde{D}}^{k+1}+\frac{\mathbf{\Gamma}_{2}^{k}}{\mu^{k}}\right),
\end{aligned}
\label{Zslove}
\end{equation}
where $\mathcal{J}(\cdot)$ is single value thresholding operator, which firstly performs singular value decomposition on $\mathbf{\widetilde{D}}^{k+1}+\mathbf{\Gamma}_{2}^{k} / \mu^{k}=\mathbf{U} \hat{\mathbf{\Sigma}} \mathbf{V}^{\top}$, and then the solution is given by $\mathbf{U} \hat{\mathbf{\Sigma}} \mathbf{V}^{\top}$, where $\hat{\Sigma}_{i i}=\max \left(0, \Sigma_{i i}-1 / \mu\right)$.

    \item $\mathbf{\widetilde{D}}$-Subproblem  is formulated as:
    \begin{equation}
\begin{aligned}
 &\min _{\mathbf{\widetilde{D}}} \beta\operatorname{Tr}\left(\mathbf{\widetilde{D} L \widetilde{D}^{T}}\right)+\langle\mathbf{\Gamma}_{1}, \mathbf{\widetilde{D}+E-D}\rangle \\
&+\langle\mathbf{\Gamma}_{2}, \mathbf{\widetilde{D}-Z}\rangle 
 +\frac{\mu}{2}\left(\|\mathbf{\widetilde{D}+E-D\|}_{F}^2+\mathbf{\|\widetilde{D}-Z\|}_{F}^{2}\right) .
\end{aligned}
\label{Dsub}
\end{equation}
Eq. (\ref{Dsub}) can be solved by setting  the first-order  derivative to zero, i.e.,  
\begin{equation}
\begin{aligned}
\mathbf{\widetilde{D}}^{k+1}=-\mu^{k}\left(\mathbf{\psi}_{1}+\mathbf{\psi}_{1}\right) /\left(2\mathbf{L}^{k}+2 \mu^{k}\mathbf{I}\right).
\end{aligned}
\label{Dslove}
\end{equation}
where $\psi_{1}=\mathbf{Z}-\mathbf{\Gamma}_{2}^{k} / \mu^{k}$, $\psi_{2}=\mathbf{D}-\mathbf{E}-\mathbf{\Gamma}_{1}^{k} / \mu^{k}$, $\mathbf{I}\in \mathbbm{R}^{n\times n}$ is an all ones matrix.
\item$\mathbf{E}$-Subproblem is represented as:
\begin{equation}
\begin{aligned}
 \min _\mathbf{E} \alpha\|\mathbf{E}\|_{1}\!+\!\langle\mathbf{\Gamma}_{1}, \mathbf{\widetilde{D}\!+\!E\!-\!D}\rangle 
 \!+\!\frac{\mu}{2}\|\mathbf{\widetilde{D}\!+\!E\!-\!D}\|_{F}^{2}.
\end{aligned}
\label{Esub}
\end{equation}
Eq. (\ref{Esub}) can be sloved by
\begin{equation}
\begin{aligned}
\mathbf{E}^{k+1}=\delta_{\alpha / \mu}\left(\mathbf{D}-\mathbf{\widetilde{D}}^{k+1}+\frac{\mathbf{\Gamma}_{1}^{k}}{\mu_{k}}\right),
\end{aligned}
\label{Eslove}
\end{equation}
where $\delta_{\alpha / \mu}(\cdot)$ is the soft-thresholding operator \cite{liu2010robust}: $\delta_{\omega}(a)=\operatorname{sgn}(a)$ for $|a| \geq \omega$ and zero otherwise.
\item The update multipliers and penalty parameter are updated by

\begin{equation}
    \left\{
        \begin{array}{l}
             \mathbf{\Gamma}_{1}^{k+1}=\mathbf{\Gamma}_{1}^{k}+\mu^{k}\left(\mathbf{\widetilde{D}}^{k+1}+\mathbf{{E}}^{k+1}-\mathbf{D}^{k+1}\right) \\
            \mathbf{\Gamma}_{2}^{k+1}=\mathbf{\Gamma}_{2}^{k}+\mu^{k}\left(\mathbf{\widetilde{D}}^{k+1}-\mathbf{Z}^{k+1}\right) \\
            \mu^{k+1}=\min \left(1.1 \mu, \mu_{\max }\right).
        \end{array}
    \right.
    \label{updatechengfa}
\end{equation}

\end{enumerate}

\subsection{Numerical Solution of  Eq. (\ref{graph})}
To solve Eq. (3), we rewrite it as
\begin{equation}
\begin{aligned}
&\min _{a_i} \sum_{j=1}^n\frac{1}{2}\left\|a_i+\frac{1}{4 r} u_i\right\|_2^2 a_{i j}\\
&\text { s.t. } {a_i}^T \mathbf{1}_n=\mathbf{1}_n, \forall i.j, 0 \leq a_{i j} \leq 1,
\end{aligned}
\label{slovegraph}
\end{equation}
where $u_{i j}=\frac{\beta}{2}\left\|x_{i}-x_{j}\right\|^{2}$. Eq. (\ref{slovegraph}) can be solved column-wisely, and the corresponding  Lagrangian function of problem (\ref{slovegraph}) regarding the $i$-th column is
\begin{equation}
\begin{aligned}
&\mathcal{L}\left(a_{i}, \varpi, \varrho\right)= \frac{1}{2}\left\|a_i+\frac{1}{4 r} u_i\right\|_2^2 a_{i j}\\
 &-\varpi\left(a_{i}^{T} \mathbf{1}_{n}-1\right)-\varrho_{i}^{T} a_{i},
\end{aligned}
\end{equation}
where $\varpi$ is a scalar and $\varrho$ is a Lagrangian coefficient vector. According to the KKT conditions \cite{boyd2004convex}, we have

\begin{equation}
    \left\{
        \begin{array}{l}
            \forall j, \quad \frac{1}{4 r} u_j+a_{i j}-\varpi-\varrho_{j}=0, \\
            \forall j, \quad \varrho_{j} \geq 0, \quad 0 \leq a_{i j} \leq 1, \\
            \forall j,  a_{i j} \varrho_{j}=0.
        \end{array}
    \right.
    \label{m and pa}
\end{equation}
After solving the KKT conditions, we have 
\begin{equation}
\begin{aligned}
a_{ j}=\left(f_j-\bar{\varrho}\right)_{+}
\end{aligned}
\label{16-18}
\end{equation}
where $\bar{\varrho}=\frac{\boldsymbol{1}^{\top} \boldsymbol{\varrho}}{n}$ and $f=\mathbf{p}-\frac{\mathbf{1 1}^T}{n} \mathbf{p}+\frac{1}{n} \mathbf{1}$, and $\bar{\varrho}$ is the root of the following equation 
\begin{equation}
\begin{aligned}
f(\bar{\varrho})=\frac{1}{n} \sum_{j=1}^{n}\left(\bar{\varrho}-f_{i j}\right)_{+}-\bar{\varrho}=0.
\end{aligned}
\label{=o}
\end{equation}
Eq. (\ref{=o}) can be solved efficiently by the Newton method
\begin{equation}
\begin{aligned}
\bar{\varrho}_{t+1}=\bar{\varrho}_{t}-\frac{f\left(\bar{\varrho}_{t}\right)}{f^{\prime}\left(\bar{\varrho}_{t}\right)},
\end{aligned}
\end{equation}
where $f^{\prime}$(X) represents the partial derivative of X.

\subsection{Numerical Solution of  Eq. (\ref{fenleimodel})}

To minimize the objective function, we opt for an iterative quasi-Newton method called Iterative Re-Weighted Least Square (IRWLS) \cite{perez2000irwls}. Initially, the objective function is approximated by its first-order Taylor expansion at the solution of the current k-th iteration, denoted by $\mathbf{\Theta}^{(k)}$:
\begin{equation}
\tilde{\ell}\left(u_i\right)=\ell\left(u_i^{(k)}\right)+\left.\frac{d \ell}{d u}\right|_{u_i^{(k)}} \frac{\left(\boldsymbol{e}_i^{(k)}\right)^{\top}}{u_i^{(k)}}\left(\boldsymbol{e}_i-\boldsymbol{e}_i^{(k)}\right)
\end{equation}

where $\boldsymbol{e}_i^{(k)}$ and $u_i^{(k)}$ are calculated using $\mathbf{\Theta}^{(k)}$ and $\mathbf{b}^{(k)}$. Subsequently, a quadratic approximation is further constructed as:
\begin{equation}
\begin{aligned}
&\bar{\ell}\left(u_i\right)  =\ell\left(u_i^{(k)}\right)+\left.\frac{d \ell\left(u_i\right)}{d u_i}\right|_{u_i^{(k)}} \frac{u_i^2-\left(u_i^{(k)}\right)^2}{2 u_i^{(k)}} \\
& =\frac{1}{2} \xi_i u_i^2+\tau,
\end{aligned}
\label{yijiejieguo}
\end{equation}
where
\begin{equation}
\xi_i=\left.\frac{1}{u_i^{(k)}} \frac{\ell\left(u_i\right)}{d u_i}\right|_{u_i^{(k)}}= \begin{cases}0 & u_i^{(k)}<\varepsilon \\ \frac{2\left(u_i^{(k)}-\varepsilon\right)}{u_i^{(k)}} & u_i^{(k)} \geq \varepsilon\end{cases}
\end{equation}
and $\tau$ is a constant term that does not rely on either  $\mathbf{\Theta}^{(k)}$  or $\mathbf{b}^{(k)}$. By combining Eq. (\ref{fenleimodel}) and (\ref{yijiejieguo}), our objective function can be rewritten as:
\begin{equation}
\begin{aligned}
&\min _{\mathbf{(\Theta, b)}} \frac{1}{2}\|\boldsymbol{\Theta}\|_F^2+\frac{1}{2} \kappa \sum_{i=1}^n \xi_i u_i^2-\nu \operatorname{tr}\left(\widetilde{\mathbf{D}}^{\top} \boldsymbol{\Theta} \boldsymbol{\Phi}\right) \\
& =\frac{1}{2}\|\boldsymbol{\Theta}\|_F^2-\nu \operatorname{tr}\left(\widetilde{\mathbf{D}}^{\top} \boldsymbol{\Theta} \boldsymbol{\Phi}\right) \\
& +\frac{1}{2} \kappa\left(\left(\widetilde{\mathbf{D}}-\boldsymbol{\Theta}^{\top} \boldsymbol{\Phi}\right) \mathbf{H}\left(\widetilde{\mathbf{D}}-\boldsymbol{\Theta}^{\top} \boldsymbol{\Phi}\right)^{\top}\right) .
\end{aligned}
\end{equation}

Here, $\mathbf{H}=\left[h_{i j}\right]_{n \times n}$, where $h_{i j}=\xi_i \delta_{i j}$, and $\delta_{i j}$ is the Kronecker's delta function. By setting the corresponding gradient to zero:
\begin{equation}
\nabla_{\boldsymbol{\Theta}}=\kappa \boldsymbol{\Phi} \mathbf{H} \boldsymbol{\Phi}^{\top} \boldsymbol{\Theta}-\kappa \boldsymbol{\Phi} \mathbf{H} \widetilde{\mathbf{D}}^{\top}+\nu \boldsymbol{\Phi} \widetilde{\mathbf{D}}^{\top}+\boldsymbol{\Theta}=\mathbf{0}
\end{equation}
the solution is obtained as
\begin{equation}
\boldsymbol{\Theta}^s=\left(\kappa \mathbf{\Phi} \mathbf{H} \boldsymbol{\Phi}^{\top}+\mathbf{I}\right)^{-1}\left(\kappa \mathbf{\Phi} \mathbf{H} \mathbf{D}^{\top}-\nu \boldsymbol{\Phi} \widetilde{\mathbf{D}}^{\top}\right)
\label{fenleislotion}
\end{equation}
Then, the solution for the next iteration, $\mathbf{\Theta}^{(k+1)}$, is obtained using a line search algorithm with $\boldsymbol{\Theta}^s$ and $\boldsymbol{\Theta}^{(k)}$. Finally, after normalizing the prediction results, we obtain the predicted label distribution. In addition, our method can also cooperate with any LDL (Label Distribution Learning) algorithm. The overall algorithm flowchart is shown in Algorithm \ref{A1}. 

\begin{algorithm}[t]
\caption{ The pseudo-code of the proposed method}\label{algorithm: }
 \textbf{Input}: 
 
$\mathbbm{T}$:  the noisy training set $\left\{\left(\mathbf{x}_i, \mathbf{d}_i\right) \mid 1 \leq i \leq n\right\}$;\\
$\alpha$, $\beta$: the trade-off parameters in the loss fuction (\ref{finnalloss});\\
$x^*$: the unseen instance to be predicted; \\   
 \textbf{Output}: \\
$\mathbf{\widetilde{D}}$, $\mathbf{E}$:  the recovered LD matrix  and the noise LD matrix; \\
$d^*$: the predicted LD for the unseen instance $x^*$ by  our approach;\\

 \textbf{Process}: 
 \begin{algorithmic}[1]
\STATE Calculate  the adaptive similarity graph  $\mathbf{A}$ by solving  Eq. (\ref{graph});
\STATE  Calculate        the graph Laplacian matrix  $\mathbf{L}$;
 \STATE Initialize the $n\times m$ ideal LD matrix $\mathbf{\widetilde{D}}$=$\mathbf{D}$;
 \STATE Initialize the $n \times m$  noise LD matrix $\mathbf{E}=0$;
 \STATE Initialize the $n \times m$  intermediate variable matrix $\mathbf{Z}$=$\mathbf{D}$;
\REPEAT
 \STATE Update $\mathbf{\widetilde{D}}$ by solving  Eq. (\ref{Dslove});
  \STATE Update $\mathbf{E}$ according to Eq. (\ref{Eslove});
 \STATE Update $\mathbf{Z}$ according to Eq. (\ref{Zslove});          
  \STATE Update the Lagrangian multipliers and penalty parameter according to Eq. 
    (\ref{updatechengfa});   
  \UNTIL{convergence}
 \RETURN $\mathbf{\widetilde{D}},\mathbf{E}$;
 \STATE Form the clean  training set $\widehat{\mathbbm{T}}=\left\{\left(x_i, \mathbf{\widetilde{D}}(:,i)\right) \mid 1 \leq i \leq n\right\}$;       
 \STATE Initialize the predictive model $\boldsymbol{\Theta}^{(0)}$, T=0;
 \REPEAT
  \STATE Calculate $\boldsymbol{\Theta}^{(s)}$ via Eq. (\ref{fenleislotion});
  \STATE Update $\boldsymbol{\Theta}^{(t+1)}$ via line searching with $\boldsymbol{\Theta}^{(t)}$ and $\boldsymbol{\Theta}^{(s)}$
  \STATE t=t+1;
  \UNTIL{convergence}\\
$\mathbf{Output}$: The predictive LD of unseen instance $\boldsymbol{\Theta(x^*)}$
 \end{algorithmic}
 \label{A1}
\end{algorithm}

\section{Experimental Results and Analyses}
\subsection{Datasets}

In this section, we present the datasets employed in our experiments to assess the performance of our proposed method. A total of 15 datasets are utilized, and their details are provided in Table \ref{tab_dataset}. The datasets encompass a broad spectrum of domains, such as biology, film, facial expression analysis, images from social media platforms, facial beauty perception, and natural scene classification. This diverse assortment of datasets enables us to evaluate the adaptability of our proposed method across various application contexts.

The first, third, and fourth datasets, M2B \cite{M2B} , SCUT-FBP \cite{SCUTxie2015scut}, and fbp5500 \cite{FP5500liang2018scut}, focus on facial beauty perception. For M2B and SCUT-FBP, the features and label distributions are processed according to \cite{ren2017sense}. For fbp5500, we utilize the ResNet \cite{he2016deep} trained by the authors to extract 512-dimensional features.

The  second datasets, RAF-ML, pertains to facial expression recognition, with each image characterized by a 2000-dimensional DBM-CNN feature and a 6-dimensional expression distribution \cite{li2019blended}. To reduce the feature dimensionality, we apply principal component analysis (PCA), resulting in 200-dimensional features.

The sixth and seventh datasets, flickr-ldl and twitter-ldl datasets \cite{yang2017learning}. These datasets comprise 10,045 and 10,700 images, respectively, annotated with 8 prevalent emotions. Both logical labels and label distributions are provided for these datasets. Image features are extracted utilizing VGGNet and subsequently dimensionality-reduced to 200 using PCA.

Seventh and Eighth datasets : These datasets are related to yeast, focusing on the budding yeast Saccharomyces cerevisiae. Each dataset represents the results of distinct biological experiments, involving a total of 2,465 yeast genes described by a phylogenetic profile vector with 24 features. The expression level of each gene at different time points is represented by the corresponding label's normalized description degree.

\begin{table}[tb]
\centering
\begin{tabular}{ccccc}
\hline
Index & Data sets   & examples & features & labels                  \\ \hline
1     & M2B   & 1240     & 250       & 5                       \\ 
2     & RAF-ML   & 4908     & 200       & 6                       \\ 
3     & SCUT-FBP  & 1500     & 300       & 5                       \\ 
4     & fbp5500  & 5500     & 512       & 5                      \\ 
5     & flickr-ldl  & 11150     & 200       & 8                      \\ 
6     & twitter-ldl & 10040     & 200       & 8                      \\ 
7     & Yeast-cdc   & 2465     & 24       & 15                      \\ 
8     & Yeast-alpha & 2465     & 24       & 18                      \\ 
9     & SBU-3DFE   & 2500     & 243       & 6                      \\ 
10    & Movie  & 7755     & 1869       & 5                       \\ 
11    & s-JAFFE     & 213      & 243      & 6                       \\ 
12    & Nature-scene    & 2000     & 294      & 9                       \\ \hline
\end{tabular}
\caption{Details of the datasets.}
\label{tab_dataset}
\end{table}

The ninth and eleventh datasets are  s-JAFFE and SBU-3DFE, are extended versions of widely-used facial expression databases, JAFFE \cite{SJAFFElyons1998coding} and BU 3DFE \cite{SBUyin20063d}, respectively. SJAFFE contains 213 grayscale images with 243-dimensional LBP features \cite{LBPahonen2006face}. Each image is scored by 60 individuals on six basic emotions, and the normalized average scores create the label distribution. Similarly, SBU 3DFE consists of 2,500 images scored by 23 individuals, resulting in a label distribution version of the dataset.

The tenth dataset is a movie genre dataset contains information about various movies and their associated genres. Features are extracted from movie metadata, such as cast, director, plot, and release year, resulting in a multi-dimensional feature vector for each movie. The label distribution is determined by calculating the proportions of each genre associated with the movie.

The last dataset is  natural scene dataset, which contains 2,000 images with inconsistent multi-label rankings. Ten human annotators ranked the images using nine possible labels. A non-linear programming process transformed the inconsistent rankings into label distributions, and a 294-dimensional feature vector was extracted for each image.

\subsection{Evaluation Metrics}

In this study, we employ a combination of six metrics to evaluate the performance of the LDL algorithms. These metrics comprise five distance-based measures and one similarity-based measure, as follows:

Chebyshev$\downarrow$
Clark$\downarrow$
Kullback-Leibler (KL)$\downarrow$
Canberra$\downarrow$
S$\phi$ren$\downarrow$
Cosine$\uparrow$
intersection$\uparrow$.
The formulas for these metrics are provided in Table 2. In these formulas, $\boldsymbol{d}$ represents the actual label distribution, and $\hat{\boldsymbol{d}}$ represents the predicted label distribution for the i-th element. Lower values indicate better performance for distance-based measures, while higher values signify better performance for similarity-based measures.

\begin{table}[]
\centering
\begin{tabular}{cc}
\hline
Measure          & Formula \\ \hline
Chebyshev $\downarrow$       &   $\operatorname{Dis}_1(\boldsymbol{d}, \hat{\boldsymbol{d}})=\max _j\left|d_j-\hat{d}_j\right|$       \\ \hline
Clark            &  $\operatorname{Dis}_2(\boldsymbol{d}, \hat{\boldsymbol{d}})=\sqrt{\sum_{j=1}^c \frac{\left(d_j-\hat{d}_j\right)^2}{\left(d_j+\hat{d}_j\right)^2}}$        \\ \hline
Canberra $\downarrow$        &   $\operatorname{Dis}_3(\boldsymbol{d}, \hat{\boldsymbol{d}})=\sum_{j=1}^c \frac{\left|d_j-\hat{d}_j\right|}{d_j+\hat{d}_j}$      \\ \hline
Kullback-Leibler$\downarrow$ &    $\operatorname{Dis}_4(\boldsymbol{d}, \hat{\boldsymbol{d}})=\sum_{j=1}^c d_j \ln \frac{d_j}{\hat{d}_j}$    \\ \hline
cosine $\uparrow$   & $\operatorname{Sim}_1(\boldsymbol{d}, \hat{\boldsymbol{d}})=\frac{\sum_{j=1}^c d_j \hat{d}_j}{\sqrt{\sum_{j=1}^c d_j^2} \sqrt{\sum_{j=1}^c \hat{d}_j^2}}$     \\ \hline

S$\phi$rensen $\downarrow$    &   $\operatorname{Dis}_5(\boldsymbol{d}, \hat{\boldsymbol{d}})=\frac{2 |\boldsymbol{d} \cap \hat{\boldsymbol{d}}|}{|\boldsymbol{d}| + |\hat{\boldsymbol{d}}|}$\\ \hline
intersection$\uparrow$ & $\operatorname{Sim}_1(\boldsymbol{d}, \hat{\boldsymbol{d}})=\sum_i \min \left(\boldsymbol{d}, \hat{\boldsymbol{d}}\right)$ \\ \hline

\end{tabular}
\caption{The distribution distance/similarity measures.}
\label{PINGJIAZHIBIAO}
\end{table}

\begin{table*}[!h]\centering
\begin{tabular}{ccccccccc}
\hline
data & algorithm & chebyshev & clark & canberra & kldist & cosine & intersection & S$\phi$rensen \\ \hline
\multirow{9}{*}{M2B} & AA-BP & 0.7020±.0888 & 1.6817±.0924 & 3.5446±.0007 & 1.7782±.0022 & 0.3246±.0003 & 0.2742±.0693 & 0.7258±.0693 \\
 & AA-KNN & 0.5596±.0338 & 1.5662±.0518 & 3.2637±.0007 & 0.7447±.0022 & 0.7160±.0003 & 0.4404±.0757 & 0.5596±.0757 \\
 & CPNN & 0.4919±.0005 & 1.6775±.0080 & 3.5298±.0148 & 0.8785±.0032 & 0.6083±.0015 & 0.4071±.0021 & 0.5929±.0021 \\
 & LDSVR & 0.5421±.0000 & 1.6846±.0009 & 3.5139±.0025 & 1.6929±.0215 & 0.5722±.0014 & 0.3925±.0014 & 0.6075±.0003 \\
 & LCLR & 0.4911±.0050 & 1.6459±.0055 & 3.3909±.0152 & 0.8058±.0323 & 0.6576±.0056 & 0.4523±.0043 & 0.5477±.0043 \\
 & LDLSF & 0.5050±.0043 & 1.6575±.0027 & 3.4292±.0084 & 0.8306±.0060 & 0.6466±.0026 & 0.4392±.0024 & 0.5608±.0024 \\
 & LDLLC & 0.4995±.0034 & 1.6489±.0028 & 3.3974±.0098 & 1.0626±.0338 & 0.6546±.0023 & 0.4499±.0024 & 0.5501±.0024 \\
 & PT-Bayes & 0.5917±.0007 & 2.0498±.0020 & 4.4067±.0034 & 1.5719±.0006 & 0.5281±.0003 & 0.4041±.0008 & 0.5959±.0008 \\
 & OURS & \textbf{0.4499±.0040} & \textbf{1.5743±.0040} & \textbf{3.2218±.0090} & \textbf{0.1592±.0085} & \textbf{0.7312±.0035} & \textbf{0.5246±.0021} & \textbf{0.4754±.0021} \\ \hline
\multirow{9}{*}{RAF} & AA-BP & 0.4080±.0006 & 1.6892±.0072 & 3.7083±.0146 & 0.2950±.0014 & 0.5491±.0010 & 0.4600±.0019 & 0.5400±.0688 \\
 & AA-KNN & 0.3573±.0021 & 1.5698±.0011 & 3.3801±.0036 & 0.0803±.0458 & 0.7137±.0016 & 0.5421±.0021 & 0.4579±.0002 \\
 & CPNN & 0.4000±.0216 & 1.1745±.0338 & 2.3595±.0463 & 0.5378±.0056 & 0.6301±.0047 & 0.6000±.0216 & 0.5612±.0974 \\
 & LDSVR & 0.4733±.0015 & 2.1164±.0009 & 4.9538±.0125 & 0.0760±.0009 & 0.5561±.0002 & 0.3334±.0010 & 0.6666±.0782 \\
 & LCLR & 0.3454±.0017 & 1.5577±.0050 & 3.3432±.0131 & 0.5786±.0047 & 0.7391±.0022 & 0.5550±.0020 & 0.4450±.0020 \\
 & LDLSF & 0.3477±.0016 & 1.6051±.0036 & 3.3787±.0102 & 0.5882±.0048 & 0.7335±.0029 & 0.5511±.0021 & 0.4489±.0021 \\
 & LDLLC & 0.4984±.0028 & 1.6526±.0021 & 3.4142±.0074 & 0.5834±.0024 & 0.6587±.0027 & 0.4490±.0022 & 0.5510±.0022 \\
 & PT-Bayes & 0.5971±.0035 & 2.1911±.0857 & 5.2174±.0362 & 0.8998±.0105 & 0.7248±.0113 & 0.4029±.0236 & 0.5971±.1247 \\
 & OURS & \textbf{0.2846±.0040} & \textbf{1.4816±.0040} & \textbf{3.0576±.0090} & \textbf{0.0211±.0085} & \textbf{0.8504±.0035} & \textbf{0.6518±.0021} & \textbf{0.3482±.0021} \\ \hline
\multirow{9}{*}{SCUT} & AA-BP & 0.3597±.1013 & 1.4320±.0014 & 2.8959±.0014 & 0.1961±.0017 & 0.7618±.0762 & 0.5443±.0748 & 0.4557±.0748 \\
 & AA-KNN & 0.6356±.0338 & 1.6880±.0967 & 3.6875±.0014 & \textbf{0.1471±.0071} & 0.6253±.0719 & 0.3644±.0634 & 0.6356±.0634 \\
 & CPNN & 0.6933±.1107 & 1.7230±.0807 & 3.7820±.0014 & 0.1903±.0021 & 0.5101±.0557 & 0.3067±.0744 & 0.6933±.0744 \\
 & LDSVR & 0.9317±.0338 & 2.6361±.0014 & 7.3787±.0014 & 0.6446±.0054 & 0.5166±.0610 & 0.3722±.0551 & 0.8031±.0551 \\
 & LCLR & 0.3501±.0040 & 1.4539±.0052 & 2.8097±.0132 & 0.5677±.0065 & 0.7434±.0032 & 0.5604±.0030 & 0.4396±.0030 \\
 & LDLSF & 0.3412±.0033 & 1.4632±.0035 & 2.8269±.0106 & 0.5622±.0046 & 0.7499±.0023 & 0.5640±.0026 & 0.4360±.0026 \\
 & LDLLC & 0.3521±.0026 & 1.4627±.0037 & 2.8317±.0095 & 0.5726±.0041 & 0.7425±.0021 & 0.5578±.0019 & 0.4422±.0019 \\
 & PT-Bayes & 0.3927±.0014 & 1.5204±.0694 & 3.0080±.0014 & 0.2213±.0069 & 0.6617±.0436 & 0.5025±.0271 & 0.4975±.0271 \\
 & OURS & \textbf{0.2468±.0029} & \textbf{1.3507±.0050} & \textbf{2.5022±.0140} & 0.1842±.0113 & \textbf{0.8469±.0023} & \textbf{0.7033±.0024} & \textbf{0.2967±.0024} \\ \hline
\multirow{9}{*}{fbp5500} & AA-BP & 0.1829±.0014 & 1.3551±.0014 & 2.4861±.0014 & 0.0988±.0017 & 0.8272±.0658 & 0.6421±.0859 & 0.3579±.0859 \\
 & AA-KNN & 0.3295±.0759 & 1.4446±.0014 & 2.7604±.0014 & 0.0981±.0016 & 0.7862±.0499 & 0.5884±.0376 & 0.4116±.0376 \\
 & CPNN & 0.3968±.0014 & 1.5044±.0940 & 2.9635±.0014 & 0.1819±.0014 & 0.6585±.0754 & 0.5022±.0745 & 0.4978±.0745 \\
 & LDSVR & 0.3270±.0014 & 1.4421±.0812 & 2.7538±.2398 & 0.0900±.0014 & 0.7922±.0612 & 0.5905±.0609 & 0.4095±.0609 \\
 & LCLR & 0.3377±.0180 & 1.4497±.0167 & 2.7787±.0556 & 0.5183±.0558 & 0.7805±.0373 & 0.5810±.0240 & 0.4190±.0240 \\
 & LDLSF & 0.3326±.0026 & 1.4497±.0689 & 2.7798±.0014 & 0.5184±.0027 & 0.7854±.2293 & 0.5861±.3806 & 0.4139±.1711 \\
 & LDLLC & 0.3334±.0016 & 1.4497±.0018 & 2.7808±.0053 & 0.5149±.0024 & 0.7832±.0012 & 0.5820±.0012 & 0.4180±.0012 \\
 & PT-Bayes & 0.3424±.0848 & 1.5953±.0014 & 3.3559±.0014 & 0.3005±.0030 & 0.6586±.0747 & 0.4508±.0475 & 0.5492±.0475 \\
 & OURS & \textbf{0.2733±.0014} & \textbf{1.3931±.0027} & \textbf{2.5892±.0072} & \textbf{0.0542±.0030} & \textbf{0.8688±.0017} & \textbf{0.6610±.0017} & \textbf{0.3390±.0017} \\ \hline
\multirow{9}{*}{flickr} & AA-BP & 0.1738±.0208 & 1.0952±.1007 & 2.5672±.3154 & 0.2797±.0017 & 0.7915±.1874 & 0.6963±.1936 & 0.3037±.0556 \\
 & AA-KNN & 0.0680±.0147 & 0.3163±.0670 & 0.7203±.1924 & 0.0300±.0017 & 0.9662±.0859 & 0.9035±.1093 & 0.0965±.0181 \\
 & CPNN & 0.8854±.0123 & 2.5295±.0570 & 7.0966±.1667 & 0.0257±.0017 & 0.7183±.0713 & 0.5095±.0937 & 0.7962±.0115 \\
 & LDSVR & 0.0637±.0022 & 0.2537±.0860 & 0.5838±.2489 & 0.0198±.0017 & 0.9759±.1113 & 0.9212±.1403 & 0.0788±.0170 \\
 & LCLR & 0.8761±.0001 & 2.5554±.0010 & 7.1757±.0029 & 7.6130±.0259 & 0.6780±.0013 & 0.4642±.0016 & 0.8042±.0002 \\
 & LDLSF & 0.8797±.0001 & 2.5562±.0005 & 7.1775±.0014 & 7.5705±.0093 & 0.6779±.0008 & 0.4634±.0008 & 0.8037±.0002 \\
 & LDLLC & 0.8761±.0001 & 2.5574±.0004 & 7.1816±.0011 & 4.3153±.3001 & 0.6750±.0004 & 0.4608±.0006 & 0.8048±.0001 \\
 & PT-Bayes & 0.6004±.0029 & 2.2813±.0083 & 6.1057±.0269 & 1.3497±.0097 & 0.4749±.0050 & 0.3143±.0026 & 0.6857±.0026 \\
 & OURS & \textbf{0.0673±.0678} & \textbf{0.2691±.2703} & \textbf{0.6256±.6275} & \textbf{0.0220±.0222} & \textbf{0.9733±.9730} & \textbf{0.9161±.9157} & \textbf{0.0839±.0843} \\ \hline
\multirow{9}{*}{twitter} & AA-BP & 0.1733±.0077 & 1.1054±.0200 & 2.5808±.0563 & 0.4290±.0189 & 0.7944±.0136 & 0.6997±.0097 & 0.3003±.0097 \\
 & AA-KNN & \textbf{0.0825±.0004} & 0.3483±.0008 & 0.7879±.0019 & 0.1228±.0004 & 0.9558±.0003 & 0.8919±.0003 & 0.1081±.0003 \\
 & CPNN & 0.0976±.0041 & 0.4117±.0271 & 0.9384±.0583 & 0.1468±.0112 & 0.9403±.0057 & 0.8715±.0079 & 0.1285±.0079 \\
 & LDSVR & 0.0873±.0005 & 0.3250±.0080 & 0.7427±.0148 & 0.1112±.0032 & 0.9583±.0015 & 0.8979±.0021 & 0.1021±.0021 \\
 & LCLR & 0.8754±.0048 & 2.6268±.0009 & 7.3826±.0025 & 5.6168±.0215 & 0.5876±.0014 & 0.3482±.0014 & 0.8177±.0003 \\
 & LDLSF & 0.8754±.0001 & 2.6267±.0007 & 7.3824±.0022 & 4.7922±.0003 & 0.5867±.0011 & 0.3482±.0012 & 0.8181±.0002 \\
 & LDLLC & 0.8755±.0000 & 2.6270±.0008 & 7.3833±.0024 & 4.7977±.0014 & 0.5861±.0012 & 0.3476±.0014 & 0.8183±.0002 \\
 & PT-Bayes & 0.5947±.0035 & 2.2634±.0085 & 6.0445±.0292 & 1.4410±.0103 & 0.4692±.0044 & 0.3190±.0030 & 0.6810±.0030 \\
 & OURS & 0.0854±.0048 & \textbf{0.3118±.0049} & \textbf{0.7175±.0144} & \textbf{0.0308±.0086} & \textbf{0.9610±.0083} & \textbf{0.9015±.0051} & \textbf{0.0985±.0051} \\ \hline
\end{tabular}
\caption{Comparison results (mean$\pm$std) measured by seven metrics. }
\label{zhushiyan1}
\end{table*}

\subsection{Inaccurate LD Matrix Generation}
To simulate the inaccurate LD, we added a controlled Gaussian noise on the ground-truth LD matrix. Specifically, we used the Matlab function randn()  to generate a random matrix of the same size as the ground-truth LD, and multiplied   the generated  random  matrix by the variance (b) and added the  mean (a) to construct the label error matrix. Then we added the error matrix to the LD matrix, and normalized the summarization as the noisy LD matrix. 

\subsection{Comparative Studies}
\subsubsection{Comparison with sota LDL algorithms}
We compare our approach with seven state-of-the-art label distribution learning approaches, using parameter configurations suggested in their respective literature:
\begin{itemize}
\item AA-BP \cite{geng2016label}: AA-BP is a structure with a three-layer network. The network outputs different units, and each output unit represents the descriptive degree of the label.
\item AA-KNN \cite{geng2016label}: For each new instance $\mathbf{x} $in AA-KNN, first find its k nearest neighbors in the training set. Then, calculate the mean of the label distribution of all k nearest neighbors as the label distribution of $\mathbf{x}$.

\item PT-Bayes \cite{geng2016label}: PT-Bayes transforms the LDL problem into a single-label learning problem, effectively converting the training samples into a set of weighted single-label samples. PT-Bayes then utilizes the Bayes classifier to address the transformed weighted single-label learning problem.

\item LCLR \cite{renLCLR}: LCLR reconstructs a new supervised label distribution with global and local label-related information. [$\lambda_1$, $\lambda_2$, $\lambda_3$, $\lambda_4$, and K are set to 0.0001, 0.001, 0.001, 0.001, and 4, respectively.]
\item LDLSF \cite{ren2019LSF}: LDLSF uses label-specific features to improve label distribution learning performance. [M are diagonal matrices in which all diagonal elements are 0.5, $\rho$ is set as $10^{-3}$]
\item LDLLC  \cite{jiaLDLLC}: LDLLC utilizes local label correlation to make prediction distributions between similar instances as close as possible.
\item CPNN \cite{geng2013facial}:  Conditional Probability Neural Network, employs a three-layer neural network structure to learn the distribution of labels.

\item LDSVR \cite{geng2015pre}:  LDSVR is to simultaneously fit a sigmoid function to each component of the label distribution using a multi-output support vector machine.
\end{itemize}
For our approach, trade-off parameters $\alpha$ and $\beta$ are set as 0.05 and 0.05 in the recover part. In the predection part, trade-off parameters $\kappa$ and $\nu$ are set  as 1 and 0.1. Table \ref{zhushiyan1} and Table \ref{zhushiyan2} present detailed experimental results comparing the algorithms using each evaluation metric. To analyze and statistically compare the performance differences between algorithms, we employ the Friedman test \cite{demvsar2006statistical}, which is a widely accepted statistical test for multiple algorithms and a specific number of datasets. For each evaluation metric, the average rank of the j-th algorithm is calculated as $R_j=\frac{1}{N} \sum_{i=1}^N r_i^j$, where $r_i^j$ represents the rank of the j-th algorithm on the i-th dataset. Subsequently, the Friedman statistics $F_F$, distributed according to the F-distribution with $(K-1)$ numerator degrees of freedom and $(K-1)(N-1)$ denominator degrees of freedom, are computed as follows:
\begin{equation}
\centering
\begin{array}{r}
F_F=\frac{(N-1) \mathcal{X}_F^2}{N(K-1)-\mathcal{X}_F^2}, \text { where } \\
\mathcal{X}_F^2=\frac{12 N}{K(K+1)}\left[\sum_{j=1}^K R_j^2-\frac{K(K+1)^2}{4}\right]
\end{array}
\label{feman test EQ}
\end{equation}
\begin{table}[!h]\centering
\centering
\begin{tabular}{lcc}
\toprule
 Evaluation metric & $F_F$ & Critical value $(\alpha=0.05)$ \\
\midrule
 Chebyshev & $29.841$ & \\
Clark & $28.9700$ & \\
KL-distance & $30.495$ & $15.51$ \\
Canberra & $32.376$ & \\
Cosine & $73.5598$ & \\
Intersection & $73.559$ & \\
S$\phi$rensen& $ 30.102$ &\\
\bottomrule
\end{tabular}

\caption{
Summary of the Friedman statistics $F_F$ 
in terms  of each evaluation  metric and
the critical value at $0.05$ significance level 
(\# comparing algorithms $K=9$, \# data sets $N=12$). }
\label{fedeman test table}
\end{table}
Table \ref{fedeman test table} summarizes the Friedman statistics $F_F$ for each evaluation metric and the corresponding critical value at significance level $\alpha=0.05$. As shown in Table \ref{fedeman test table}, the indicators of all evaluation methods exceed the critical value, i.e., the hypothesis that all algorithms perform the same is rejected, indicating that the performance of the algorithms is significantly different. 

To further distinguish the performance among the comparing algorithms, a post-hoc test is necessary at this stage. We employ the Bonferroni-Dunn test \cite{dunn1961multiple}. LSag is treated as the control algorithm, and the difference between the average ranks of IDI-LDL and one comparing algorithm is compared with the critical difference (CD). If their difference is larger than one CD (CD=2.994 with K = 9 and N = 12 at a significance level of $\alpha=0.05$), the performance of LSag is deemed to be significantly different from that of the comparing algorithm.

\begin{table*}[!h]\centering
\begin{tabular}{ccccccccc}
\hline
 &  & chebyshev & clark & canberra & kldist & cosine & intersection & S$\phi$rensen\\ \hline
\multirow{9}{*}{alpha} & AA-BP & 0.0434±.0001 & 0.9215±.0005 & 3.1670±.0014 & 0.0885±.0093 & 0.9370±.0008 & 0.8393±.0002 & 0.1607±.0006 \\
 & AA-KNN & 0.0275±.0401 & 0.5602±.0232 & 1.9409±.0964 & 0.0336±.0349 & 0.9691±.0252 & 0.8955±.0811 & 0.1045±.0028 \\
 & CPNN & 0.0724±.0025 & 0.8629±.0018 & 3.2067±.0088 & 0.0933±.0045 & 0.9028±.0016 & 0.8121±.0015 & 0.1879±.0057 \\
 & LDSVR & 0.0294±.0029 & 0.5970±.0041 & 1.9463±.0157 & 0.0390±.0096 & 0.9652±.0028 & 0.8947±.0021 & 0.1053±.0006 \\
 & LCLR & 0.0151±.0016 & 0.2412±.0011 & 0.7946±.0031 & 0.0070±.0036 & 0.9930±.0013 & 0.9560±.0002 & 0.0440±.0054 \\
 & LDLSF & 0.0151±.0023 & 0.2413±.0032 & 0.7953±.0109 & 0.0070±.0070 & 0.9930±.0093 & 0.9560±.0032 & 0.0440±.0033 \\
 & LDLLC & 0.0145±.0029 & 0.2252±.0033 & \textbf{0.7344±.0083} & 0.4233±.0054 & \textbf{0.9939±.0019} & 0.9584±.0018 & \textbf{0.0406±.0083} \\
 & PT-Bayes & 0.3875±.0015 & 2.0231±.0010 & 7.6577±.0039 & 0.6384±.0048 & 0.4843±.0016 & 0.5185±.0009 & 0.4815±.0018 \\
 & OURS & \textbf{0.0143±.0006} & \textbf{0.2201±.0028} & 0.7446±.0057 & \textbf{0.0063±.0006} & 0.9938±.0006 & \textbf{0.9590±.0007} & 0.0411±.0080 \\ \hline
\multirow{9}{*}{cdc} & AA-BP & 0.0669±.0024 & 1.0440±.0036 & 2.5764±.0079 & 0.1639±.0069 & 0.9139±.0021 & 0.8433±.0017 & 0.1567±.0028 \\
 & AA-KNN & 0.0448±.0106 & 0.6288±.0276 & 2.0304±.0790 & 0.0543±.0199 & 0.9484±.0213 & 0.8629±.0130 & 0.1371±.0076 \\
 & CPNN & 0.0620±.0058 & 1.4701±.0024 & 4.6585±.0080 & 0.2909±.0207 & 0.8524±.0011 & 0.7283±.0013 & 0.2717±.0030 \\
 & LDSVR & 0.0232±.0020 & 0.2772±.0036 & 0.8296±.0144 & 0.0094±.0091 & 0.9910±.0038 & 0.9459±.0020 & 0.0541±.0019 \\
 & LCLR & 0.0168±.0020 & 0.2237±.0027 & 0.6752±.0085 & 0.0074±.0012 & 0.9928±.0011 & 0.9555±.0011 & 0.0445±.0021 \\
 & LDLSF & 0.0168±.0020 & 0.2236±.0036 & 0.6748±.0144 & 0.0074±.0091 & 0.9928±.0038 & 0.9555±.0020 & 0.0445±.0211 \\
 & LDLLC & 0.0164±.0024 & \textbf{0.2147±.0036} & \textbf{0.6435±.0079} & 0.9225±.0069 & 0.9933±.0021 & 0.9576±.0017 & \textbf{0.0424±.0028} \\
 & PT-Bayes & 0.2408±.0042 & 1.9271±.0135 & 6.8458±.0342 & 0.5819±.0023 & 0.6639±.0026 & 0.5617±.0044 & 0.4383±.0029 \\
 & OURS & \textbf{0.0161±.0028} & 0.2163±.0076 & 0.6493±.0211 & \textbf{0.0073±.0030} & \textbf{0.9942±.0019} & \textbf{0.9578±.0021} & 0.0428±.0019 \\ \hline
\multirow{9}{*}{sja} & AA-BP & 0.4391±.0001 & 1.3493±.0004 & 2.7821±.0011 & 0.7564±61.4300 & 0.5295±.0004 & 0.5365±.0006 & 0.4635±.0001 \\
 & AA-KNN & 0.1646±.0001 & 0.5756±.0005 & 1.0271±.0014 & 0.1006±.0093 & 0.9024±.0008 & 0.8354±.0008 & 0.1926±.0002 \\
 & CPNN & 0.1986±.0040 & 0.7276±.0040 & 1.3999±.0090 & 0.1796±.0085 & 0.8364±.0035 & 0.7681±.0021 & 0.2319±.0021 \\
 & LDSVR & 0.1716±.0795 & 0.6218±.0031 & 1.1184±.0042 & 0.1175±.0061 & 0.8917±.0016 & 0.8284±.0015 & 0.1716±.0012 \\
 & LCLR & 0.1277±.0039 & 0.4430±.0057 & 0.9331±.0080 & 0.0792±.0018 & 0.9248±.0016 & 0.8400±.0759 & 0.1600±.0748 \\
 & LDLSF & 0.1240±.0051 & 0.4693±.0112 & 0.9814±.0232 & 0.0754±.0039 & 0.9309±.0037 & 0.8381±.0043 & 0.1619±.0043 \\
 & LDLLC & 0.1129±.0227 & 0.4690±.0731 & 0.9822±.0878 & 4.4745±.0565 & 0.9308±.0357 & 0.8380±.0256 & 0.1620±.1174 \\
 & PT-Bayes & 0.2714±.0584 & 0.7375±.0918 & 1.4657±.0192 & 0.2256±.0168 & 0.7842±.0190 & 0.7286±.0557 & 0.2714±.0782 \\
 & OURS & \textbf{0.0940±.0227} & \textbf{0.1893±.0731} & \textbf{0.2903±.0878} & \textbf{0.0342±.0565} & \textbf{0.9721±.0357} & \textbf{0.9060±.0256} & \textbf{0.0940±.0002} \\ \hline
\multirow{9}{*}{SBU} & AA-BP & 0.3012±.0040 & 1.1851±.0080 & 2.7515±.0187 & 0.6154±1.5501 & 0.5961±.0037 & 0.5337±.0036 & 0.4663±.0036 \\
 & AA-KNN & 0.2419±.0029 & 0.6145±.0033 & 1.3260±.0083 & 0.3355±.0054 & 0.8456±.0019 & 0.6884±.0018 & 0.2546±.0018 \\
 & CPNN & 0.2689±.0053 & 0.8444±.0277 & 1.9320±.0656 & 0.3131±.0149 & 0.7359±.0077 & 0.6558±.0089 & 0.3442±.0089 \\
 & LDSVR & 0.2324±.0279 & 1.1012±.0238 & 2.4310±.0621 & 0.4319±.1620 & 0.7189±.0093 & 0.6033±.0076 & 0.3967±.0112 \\
 & LCLR & 0.1345±.0027 & 0.4134±.0068 & 0.9043±.0160 & 0.0848±.0021 & 0.9179±.0022 & 0.8384±.0028 & 0.1616±.0028 \\
 & LDLSF & 0.1383±.0000 & 0.4112±.0008 & 0.8998±.0024 & 0.0840±.0014 & 0.9186±.0012 & 0.8392±.0014 & 0.1608±.0002 \\
 & LDLLC & 0.1400±.0023 & 0.4139±.0032 & 0.9045±.0109 & 0.0859±.0070 & 0.9169±.0093 & 0.8381±.0032 & 0.1619±.0033 \\
 & PT-Bayes & 0.3044±.0029 & 0.8913±.0033 & 1.9535±.0083 & 0.4238±.0054 & 0.6885±.0019 & 0.6276±.0018 & 0.3724±.0083 \\
 & OURS & \textbf{0.1270±.0001} & \textbf{0.3919±.0007} & \textbf{0.8499±.0022} & \textbf{0.0672±.0003} & \textbf{0.9288±.0011} & \textbf{0.8485±.0012} & \textbf{0.1515±.0002} \\ \hline
\multirow{9}{*}{MOVIE} & AA-BP & 0.1743±.0024 & 0.7322±.0036 & 1.3920±.0079 & 0.4005±.0069 & 0.8671±.0021 & 0.7569±.0017 & 0.2431±.0083 \\
 & AA-KNN & 0.1695±.0007 & 0.7123±.0022 & 1.3483±.0045 & 0.3992±.0018 & 0.8775±.0007 & 0.7596±.0008 & 0.2404±.0008 \\
 & CPNN & 0.1786±.0030 & 0.7414±.0088 & 1.4142±.0182 & 0.4299±.0108 & 0.8629±.0053 & 0.7439±.0046 & 0.2561±.0046 \\
 & LDSVR & 0.1807±.0016 & 0.7508±.0035 & 1.4321±.0077 & 0.4411±.0038 & 0.8593±.0020 & 0.7398±.0018 & 0.2602±.0018 \\
 & LCLR & 0.1654±.0042 & 0.7093±.0135 & 1.3432±.0342 & 0.1683±.0023 & 0.8827±.0026 & 0.7615±.0044 & 0.2385±.0029 \\
 & LDLSF & 0.1735±.0028 & 0.7322±.0076 & 1.3915±.0211 & 0.1829±.0030 & 0.8704±.0019 & 0.7501±.0021 & 0.2499±.0054 \\
 & LDLLC & 0.1817±.0019 & 0.7552±.0063 & 1.4398±.0174 & 0.1976±.0135 & 0.8581±.0017 & 0.7386±.0023 & 0.2614±.0019 \\
 & PT-Bayes & 0.1850±.0011 & 0.7627±.0025 & 1.4609±.0061 & 0.4506±.0025 & 0.8564±.0009 & 0.7357±.0011 & 0.2643±.0011 \\
 & OURS & \textbf{0.1338±.0040} & \textbf{0.6105±.0080} & \textbf{1.1399±.0187} & \textbf{0.1464±1.5501} & \textbf{0.9222±.0037} & \textbf{0.8077±.0036} & \textbf{0.1923±.0036} \\ \hline
\multirow{9}{*}{NATURE} & AA-BP & 0.4040±.0104 & 2.5361±.0126 & 7.1280±.0630 & 3.9933±.1458 & 0.5036±.0254 & 0.3678±.0175 & 0.6322±.0175 \\
 & AA-KNN & 0.3566±.0059 & 2.4758±.0096 & 6.9364±.0399 & 3.7662±.0269 & 0.6313±.0038 & 0.3948±.0045 & 0.6052±.0045 \\
 & CPNN & 0.3818±.0064 & 2.5189±.0106 & 7.1490±.0474 & 4.1641±.0636 & 0.5343±.0166 & 0.3388±.0087 & 0.6612±.0087 \\
 & LDSVR & 0.3642±.0061 & 2.4766±.0074 & 6.9645±.0277 & 3.9580±.0179 & 0.5798±.0032 & 0.3679±.0028 & 0.6321±.0028 \\
 & LCLR & 0.3583±.0036 & 2.4808±.0155 & 6.8452±.0285 & 1.1300±.0068 & 0.6766±.0047 & 0.4662±.0055 & 0.5338±.0055 \\
 & LDLSF & 0.3744±.0090 & 2.5772±.0234 & 7.1752±.0513 & 1.7217±.0090 & 0.6083±.0101 & 0.4530±.0100 & 0.5470±.0100 \\
 & LDLLC & 0.6816±.0027 & 2.8773±.0068 & 8.4701±.0160 & 2.8957±.0021 & 0.4247±.0022 & 0.3069±.0028 & 0.6931±.0028 \\
 & PT-Bayes & 0.4285±.0076 & 2.5436±.0073 & 7.2272±.0297 & 3.9983±.0667 & 0.5423±.0064 & 0.3399±.0052 & 0.6601±.0052 \\
 & OURS & \textbf{0.3177±.0030} & \textbf{2.4525±.0050} & \textbf{6.7557±.0208} & \textbf{1.1300±.0067} & \textbf{0.7158±.0020} & \textbf{0.4749±.0023} & \textbf{0.5251±.0023} \\ \hline
\end{tabular}
\caption{Comparison results (mean$\pm$std) measured by seven metrics. }
\label{zhushiyan2}
\end{table*}
\begin{figure*}[!h]
\centering
\includegraphics[width=1\textwidth]{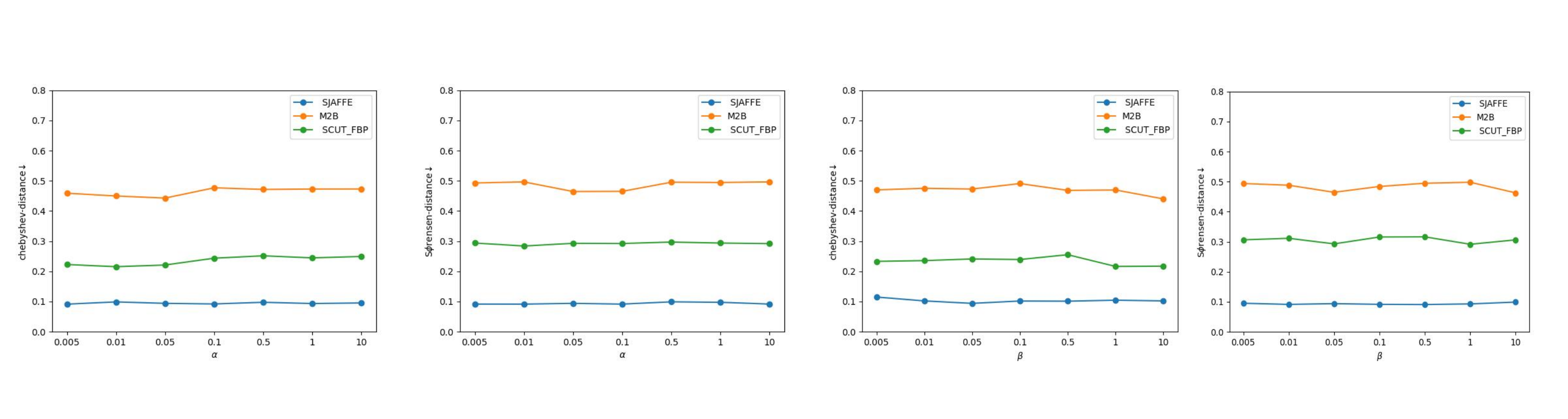} 
\caption{Performance of the proposed method  as the trade-off parameter $\alpha$ and $\beta$ vary on different data sets..}
\label{fig_real_inaccurate_reLD} 
\label{canshu1}
\end{figure*}

\begin{figure*}[!h]
\centering
\includegraphics[width=1\textwidth]{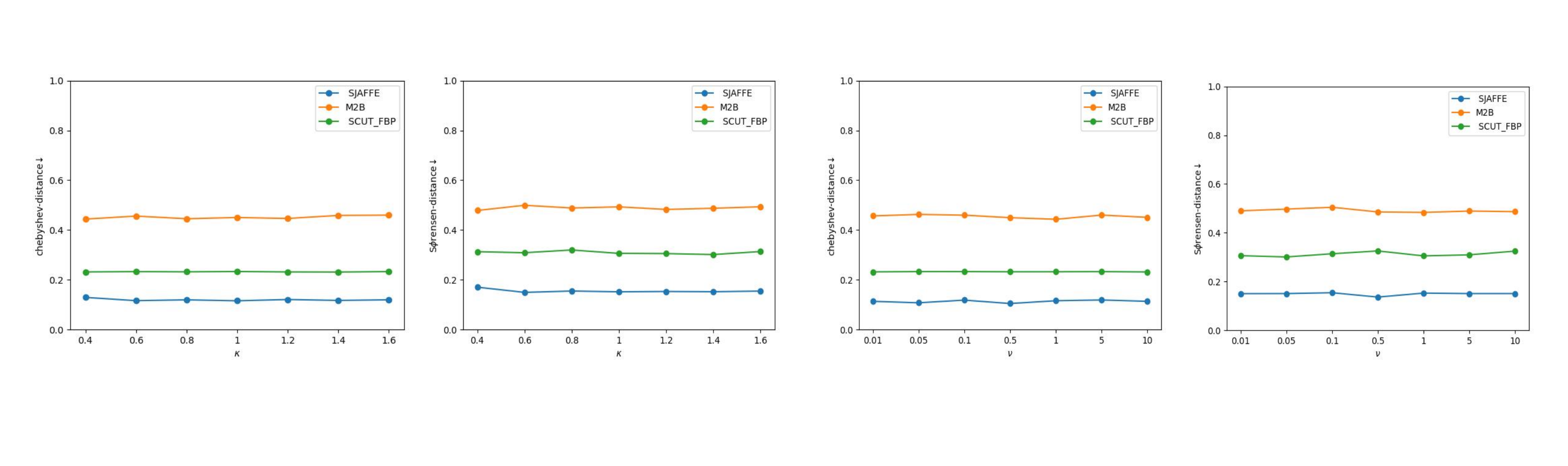} 
\caption{Performance of the proposed method  as the trade-off parameter $\kappa$ and $\nu$ vary on different data sets.}
\label{fig_real_inaccurate_reLD} 
\label{canshu2}
\end{figure*}

\begin{figure*}
    \centering
    
    \includegraphics[width=0.95\linewidth]{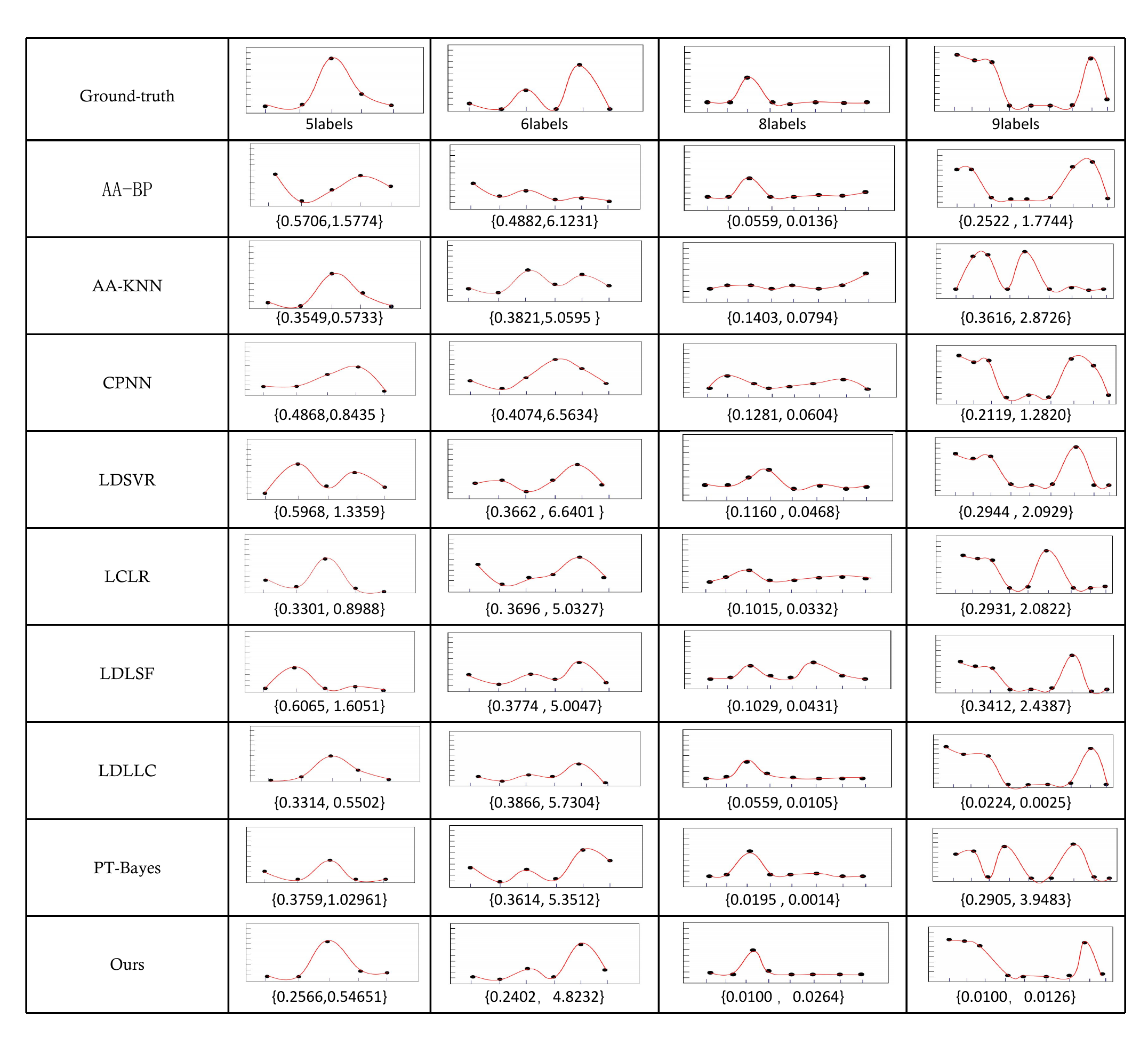}
    \caption{ Typical examples of the real and predicted label distributions, which measured by Chebyshev and KL.}
    \label{keshihua}
\end{figure*}

\begin{figure*}[!h]
\centering
\includegraphics[width=0.8\textwidth]{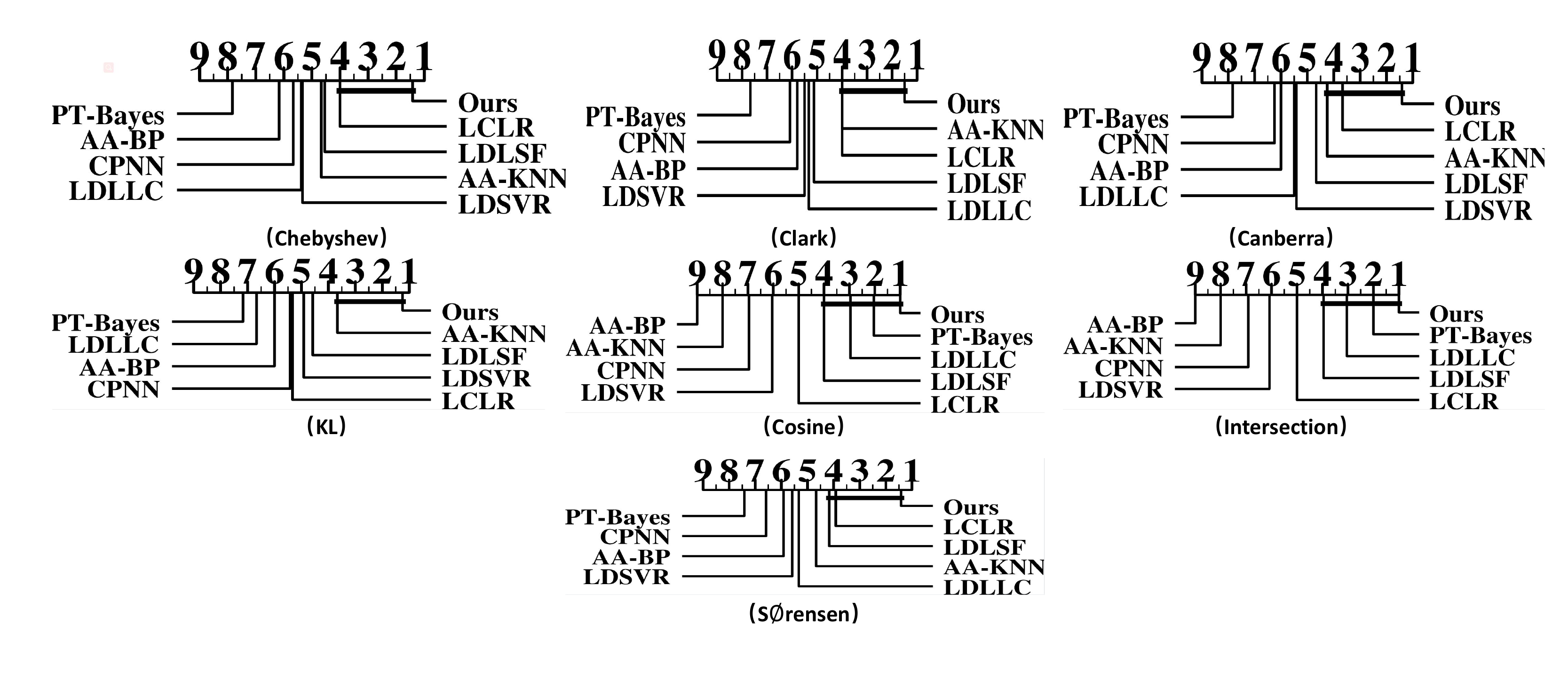} 
\caption{ Comparison of LSag (control algorithm) against other comparing algorithms with the Bonferroni-Dunn test. Algorithms not connected with
LSag  are considered to have significantly different performance from the control algorithm (significance level $\alpha$= 0.05).}
\label{fig_real_inaccurate_reLD} 
\label{CD}
\end{figure*}

Fig. \ref{CD} presents the CD diagrams \cite{demvsar2006statistical} for each evaluation metric. In each sub-figure, the average rank of each comparing algorithm is marked along the axis with lower ranks to the right, and a thick line connects LSag and any comparing algorithm if the difference between their average ranks is less than one CD. Additionally, the average ranking of the compared algorithms across 12 datasets is shown in Table \ref{rank_table}.  Based on the above results, observations can be made as follows:
\begin{itemize}
    \item In 99.21\% of the cases, our algorithm achieved the best results. This can be attributed to the fact that traditional algorithms such as AA-BP, AA-KNN, LDSVR, CPNN, and PT-Bayes do not consider the presence of noise in the label distribution, which can lead to incorrect guidance for classifier learning. 
    \item On the other hand, although LDLLC, LDL-SF, and LCLR take label correlation into account, they also fail to address the noise issue in the label distribution. As a result, the models they learn underperform.
    \item Our algorithm consistently achieves the highest average ranking, as it takes into account the noise present in the label distribution, while other methods do not.
    \item Under the Chebyshev$\downarrow$ metric, our algorithm is significantly better than all other algorithms, except for LCLR. Similar results can be observed under the Canberra$\downarrow$ and S$\phi$ren$\downarrow$ metrics. For the Clark$\downarrow$ metric, our algorithm outperforms all algorithms except for AA-KNN, and this performance is also replicated under the Kullback-Leibler (KL)$\downarrow$ and Cosine$\uparrow$ metrics. Under the intersection$\uparrow$ metric, our algorithm is significantly better than LCLR, AA-BP,  AA-KNN, CPNN and LDSVR.
\end{itemize}

\begin{table*}[]\centering
\begin{tabular}{cccccccccc}
\hline
 & AA-BP & AA-KNN & CPNN & LDSVR & LCLR & LDLSF & LDLLC & PT-Bayes & Ours \\ \hline
Chebyshev & 6.17 & 4.67 & 5.67 & 5.33 & 4.00 & 4.54 & 5.38 & 7.83 & \textbf{1.42} \\
Clark & 5.79 & 4.00 & 6.08 & 5.50 & 4.00 & 5.13 & 5.33 & 7.67 & \textbf{1.50} \\
Canberra & 6.00 & 4.25 & 6.25 & 5.42 & 3.67 & 4.67 & 5.50 & 7.83 & \textbf{1.42} \\
kl & 6.00 & 3.67 & 5.42 & 4.92 & 5.33 & 4.58 & 6.67 & 7.17 & \textbf{1.25} \\
Cosine & 9.00 & 8.00 & 6.96 & 6.04 & 5.00 & 4.00 & 2.96 & 2.04 & \textbf{1.00} \\
Intersection & 9.00 & 8.00 & 6.92 & 6.08 & 5.00 & 3.96 & 3.04 & 2.00 & \textbf{1.00} \\
Sorensen & 5.92 & 4.67 & 6.58 & 5.58 & 3.92 & 4.17 & 5.33 & 7.42 & \textbf{1.42} \\ \hline
\end{tabular}
\caption{Average ranking of algorithms across 12 datasets under different metrics.}
\label{rank_table}
\end{table*}

\begin{table*}[!h]\scriptsize 
\setlength{\tabcolsep}{1mm}
\centering
\begin{tabular}{cccccccccccccccc}
\hline
\multicolumn{16}{c}{Chebyshev$\downarrow$}                                                                                                                                                                             \\ \hline
\multicolumn{1}{l}{}        & \multicolumn{3}{c|}{$\mathcal{F}$=AABP}                                       & \multicolumn{3}{c|}{$\mathcal{F}$=AA-KNN}                                     & \multicolumn{3}{c|}{$\mathcal{F}$=CPNN}                                       & \multicolumn{3}{c|}{$\mathcal{F}$=LDSVR}                                      & \multicolumn{3}{c}{$\mathcal{F}$=PT-Bayes}                \\ \hline
\multicolumn{1}{l}{Dataset} & $\mathcal{F}$(G) & $\mathcal{F}$-LSag & \multicolumn{1}{c|}{$\mathcal{F}$(I)} & $\mathcal{F}$(G) & $\mathcal{F}$-LSag & \multicolumn{1}{c|}{$\mathcal{F}$(I)} & $\mathcal{F}$(G) & $\mathcal{F}$-LSag & \multicolumn{1}{c|}{$\mathcal{F}$(I)} & $\mathcal{F}$(G) & $\mathcal{F}$-LSag & \multicolumn{1}{c|}{$\mathcal{F}$(I)} & $\mathcal{F}$(GT) & $\mathcal{F}$-LSag & $\mathcal{F}$(I) \\
Yeast-alpha                 & 0.0750           & \textbf{0.0843}    & \multicolumn{1}{c|}{0.1207}           & 0.0142           & \textbf{0.0217}    & \multicolumn{1}{c|}{0.0275}           & 0.0126           & \textbf{0.0101}    & \multicolumn{1}{c|}{0.0724}           & 0.0150           & \textbf{0.0117}    & \multicolumn{1}{c|}{0.0193}           & 0.1269            & \textbf{0.0767}    & 0.3875           \\
Yeast-cdc                   & 0.0696           & \textbf{0.0303}    & \multicolumn{1}{c|}{0.1147}           & 0.0238           & \textbf{0.0242}    & \multicolumn{1}{c|}{0.0448}           & 0.0249           & \textbf{0.0198}    & \multicolumn{1}{c|}{0.0620}           & 0.0201           & \textbf{0.0204}    & \multicolumn{1}{c|}{0.0232}           & 0.1422            & \textbf{0.1675}    & 0.2408           \\

s-JAFFE                     & 0.0839           & \textbf{0.0796}    & \multicolumn{1}{c|}{0.4391}           & 0.0991           & \textbf{0.0883}    & \multicolumn{1}{c|}{0.1646}           & 0.0302           & \textbf{0.0667}    & \multicolumn{1}{c|}{0.1986}           & 0.2197           & \textbf{0.0828}    & \multicolumn{1}{c|}{0.1716}           & 0.1512            & \textbf{0.0828}    & 0.2714           \\
SBU 3DFE                    & 0.2981           & \textbf{0.1679}    & \multicolumn{1}{c|}{0.3012}           & 0.2376           & \textbf{0.1715}    & \multicolumn{1}{c|}{0.2419}           & 0.0732           & \textbf{0.1475}    & \multicolumn{1}{c|}{0.2689}           & 0.1983           & \textbf{0.2158}    & \multicolumn{1}{c|}{0.2324}           & 0.0828            & \textbf{0.2179}    & 0.3044           \\ \hline
Average rank                &                  & 1                  & \multicolumn{1}{c|}{2}                &                  & 1                  & \multicolumn{1}{c|}{2}                &                  & 1                  & \multicolumn{1}{c|}{2}                &                  & 1                  & \multicolumn{1}{c|}{2}                &                   & 1                  & 2                \\ \hline
\multicolumn{16}{c}{clark$\downarrow$}                                                                                                                                                                                                                                                                                                                                                                                  \\ \hline
\multicolumn{1}{l}{}        & \multicolumn{3}{c|}{$\mathcal{F}$=AABP}                                       & \multicolumn{3}{c|}{$\mathcal{F}$=AA-KNN}                                     & \multicolumn{3}{c|}{$\mathcal{F}$=CPNN}                                       & \multicolumn{3}{c|}{$\mathcal{F}$=LDSVR}                                      & \multicolumn{3}{c}{$\mathcal{F}$=PT-Bayes}                \\\hline
\multicolumn{1}{l}{Dataset} & $\mathcal{F}$(G) & $\mathcal{F}$-LSag & \multicolumn{1}{c|}{$\mathcal{F}$(I)} & $\mathcal{F}$(G) & $\mathcal{F}$-LSag & \multicolumn{1}{c|}{$\mathcal{F}$(I)} & $\mathcal{F}$(G) & $\mathcal{F}$-LSag & \multicolumn{1}{c|}{$\mathcal{F}$(I)} & $\mathcal{F}$(G) & $\mathcal{F}$-LSag & \multicolumn{1}{c|}{$\mathcal{F}$(I)} & $\mathcal{F}$(GT) & $\mathcal{F}$-LSag & $\mathcal{F}$(I) \\\hline
Yeast-alpha                 & 1.3956           & \textbf{1.7023}    & \multicolumn{1}{c|}{1.8428}           & 0.2132           & \textbf{0.4554}    & \multicolumn{1}{c|}{0.5602}           & 0.1927           & \textbf{0.1816}    & \multicolumn{1}{c|}{0.8629}           & 0.2486           & \textbf{0.1765}    & \multicolumn{1}{c|}{0.4174}           & 1.9051            & \textbf{0.0767}    & 0.3875           \\
Yeast-cdc                   & 1.5694           & \textbf{0.4751}    & \multicolumn{1}{c|}{1.6540}           & 0.2892           & \textbf{0.5043}    & \multicolumn{1}{c|}{0.6288}           & 0.3007           & \textbf{0.2488}    & \multicolumn{1}{c|}{1.4701}           & 0.2539           & \textbf{0.2424}    & \multicolumn{1}{c|}{0.2772}           & 1.4234            & \textbf{1.4526}    & 1.9271           \\

s-JAFFE                     & 1.3260           & \textbf{0.4982}    & \multicolumn{1}{c|}{1.3493}           & 0.6934           & \textbf{0.4760}    & \multicolumn{1}{c|}{0.5756}           & 0.3394           & \textbf{0.3026}    & \multicolumn{1}{c|}{0.7276}           & 0.6493           & \textbf{0.3650}    & \multicolumn{1}{c|}{0.6218}           & 0.3650            & \textbf{0.3650}    & 0.7375           \\
SBU 3DFE                    & 0.6008           & \textbf{0.9449}    & \multicolumn{1}{c|}{1.1851}           & 0.3962           & \textbf{0.4902}    & \multicolumn{1}{c|}{0.6145}           & 0.4644           & \textbf{0.5130}    & \multicolumn{1}{c|}{0.8444}           & 0.5602           & \textbf{0.6509}    & \multicolumn{1}{c|}{1.1012}           & 0.6485            & \textbf{0.6485}    & 0.8913           \\\hline
Average rank                &                  & 1                  & \multicolumn{1}{c|}{2}                &                  & 1                  & \multicolumn{1}{c|}{2}                &                  & 1                  & \multicolumn{1}{c|}{2}                &                  & 1                  & \multicolumn{1}{c|}{2}                &                   & 1                  & 2                \\\hline
\multicolumn{16}{c}{cosine$\uparrow$}                                                                                                                                                                                                                                                                                                                                                                                   \\\hline
\multicolumn{1}{l}{}        & \multicolumn{3}{c|}{$\mathcal{F}$=AABP}                                       & \multicolumn{3}{c|}{$\mathcal{F}$=AA-KNN}                                     & \multicolumn{3}{c|}{$\mathcal{F}$=CPNN}                                       & \multicolumn{3}{c|}{$\mathcal{F}$=LDSVR}                                      & \multicolumn{3}{c}{$\mathcal{F}$=PT-Bayes}                \\\hline
\multicolumn{1}{l}{Dataset} & $\mathcal{F}$(G) & $\mathcal{F}$-LSag & \multicolumn{1}{c|}{$\mathcal{F}$(I)} & $\mathcal{F}$(G) & $\mathcal{F}$-LSag & \multicolumn{1}{c|}{$\mathcal{F}$(I)} & $\mathcal{F}$(G) & $\mathcal{F}$-LSag & \multicolumn{1}{c|}{$\mathcal{F}$(I)} & $\mathcal{F}$(G) & $\mathcal{F}$-LSag & \multicolumn{1}{c|}{$\mathcal{F}$(I)} & $\mathcal{F}$(GT) & $\mathcal{F}$-LSag & $\mathcal{F}$(I) \\\hline
Yeast-alpha                 & 0.8489           & \textbf{0.8195}    & \multicolumn{1}{c|}{0.7641}           & 0.9949           & \textbf{0.9965}    & \multicolumn{1}{c|}{0.9691}           & 0.9958           & \textbf{0.9963}    & \multicolumn{1}{c|}{0.9028}           & 0.9937           & \textbf{0.9965}    & \multicolumn{1}{c|}{0.9814}           & 0.7442            & \textbf{0.8520}    & 0.4843           \\
Yeast-cdc                   & 0.8267           & \textbf{0.9717}    & \multicolumn{1}{c|}{0.7674}           & 0.9902           & \textbf{0.9933}    & \multicolumn{1}{c|}{0.9484}           & 0.9894           & \textbf{0.9929}    & \multicolumn{1}{c|}{0.8524}           & 0.9925           & \textbf{0.9933}    & \multicolumn{1}{c|}{0.9910}           & 0.7867            & \textbf{0.8703}    & 0.6639           \\

s-JAFFE                     & 0.6994           & \textbf{0.9439}    & \multicolumn{1}{c|}{0.5295}           & 0.8110           & \textbf{0.9726}    & \multicolumn{1}{c|}{0.9024}           & 0.9759           & \textbf{0.9737}    & \multicolumn{1}{c|}{0.8364}           & 0.8563           & \textbf{0.9710}    & \multicolumn{1}{c|}{0.8917}           & 0.9710            & \textbf{0.9710}    & 0.7842           \\
SBU 3DFE                    & 0.8251           & \textbf{0.8927}    & \multicolumn{1}{c|}{0.5961}           & 0.9572           & \textbf{0.9139}    & \multicolumn{1}{c|}{0.8456}           & 0.9265           & \textbf{0.9222}    & \multicolumn{1}{c|}{0.7359}           & 0.8832           & \textbf{0.8456}    & \multicolumn{1}{c|}{0.7189}           & 0.8454            & \textbf{0.8454}    & 0.6885           \\\hline
Average rank                &                  & 1                  & \multicolumn{1}{c|}{2}                &                  & 1                  & \multicolumn{1}{c|}{2}                &                  & 1                  & \multicolumn{1}{c|}{2}                &                  & 1                  & \multicolumn{1}{c|}{2}                &                   & 1                  & 2                \\\hline
\multicolumn{16}{c}{S$\phi$rensen$\downarrow$}                                                                                                                                                                                                                                                                                                                                                                           \\\hline
\multicolumn{1}{l}{}        & \multicolumn{3}{c|}{$\mathcal{F}$=AABP}                                       & \multicolumn{3}{c|}{$\mathcal{F}$=AA-KNN}                                     & \multicolumn{3}{c|}{$\mathcal{F}$=CPNN}                                       & \multicolumn{3}{c|}{$\mathcal{F}$=LDSVR}                                      & \multicolumn{3}{c}{$\mathcal{F}$=PT-Bayes}                \\\hline
\multicolumn{1}{l}{Dataset} & $\mathcal{F}$(G) & $\mathcal{F}$-LSag & \multicolumn{1}{c|}{$\mathcal{F}$(I)} & $\mathcal{F}$(G) & $\mathcal{F}$-LSag & \multicolumn{1}{c|}{$\mathcal{F}$(I)} & $\mathcal{F}$(G) & $\mathcal{F}$-LSag & \multicolumn{1}{c|}{$\mathcal{F}$(I)} & $\mathcal{F}$(G) & $\mathcal{F}$-LSag & \multicolumn{1}{c|}{$\mathcal{F}$(I)} & $\mathcal{F}$(GT) & $\mathcal{F}$-LSag & $\mathcal{F}$(I) \\\hline
Yeast-alpha                 & 0.2340           & \textbf{0.2946}    & \multicolumn{1}{c|}{0.3569}           & 0.0412           & \textbf{0.1010}    & \multicolumn{1}{c|}{0.1045}           & 0.0370           & \textbf{0.0353}    & \multicolumn{1}{c|}{0.1879}           & 0.0438           & \textbf{0.0327}    & \multicolumn{1}{c|}{0.0811}           & 0.3447            & \textbf{0.2399}    & 0.4815           \\
Yeast-cdc                   & 0.2898           & \textbf{0.0889}    & \multicolumn{1}{c|}{0.3179}           & 0.0526           & \textbf{0.1248}    & \multicolumn{1}{c|}{0.1371}           & 0.0510           & \textbf{0.0406}    & \multicolumn{1}{c|}{0.2717}           & 0.0511           & \textbf{0.0442}    & \multicolumn{1}{c|}{0.0541}           & 0.3070            & \textbf{0.2354}    & 0.4383           \\

s-JAFFE                     & 0.4115           & \textbf{0.1699}    & \multicolumn{1}{c|}{0.4635}           & 0.2376           & \textbf{0.1646}    & \multicolumn{1}{c|}{0.1926}           & 0.0899           & \textbf{0.0912}    & \multicolumn{1}{c|}{0.2319}           & 0.2017           & \textbf{0.0958}    & \multicolumn{1}{c|}{0.1716}           & 0.0958            & \textbf{0.0958}    & 0.2714           \\
SBU 3DFE                    & 0.2175           & \textbf{0.2562}    & \multicolumn{1}{c|}{0.4663}           & 0.1256           & \textbf{0.1784}    & \multicolumn{1}{c|}{0.2546}           & 0.1662           & \textbf{0.1862}    & \multicolumn{1}{c|}{0.3442}           & 0.2214           & \textbf{0.2619}    & \multicolumn{1}{c|}{0.3967}           & 0.2584            & \textbf{0.2584}    & 0.3724           \\\hline
Average rank                &                  & 1                  & \multicolumn{1}{c|}{2}                &                  & 1                  & \multicolumn{1}{c|}{2}                &                  & 1                  & \multicolumn{1}{c|}{2}                &                  & 1                  & \multicolumn{1}{c|}{2}                &                   & 1                  & 2    \\ \hline      
\end{tabular}
\caption{Prediction results  measured by  (Chebyshev $\downarrow$, Clark$\downarrow$,  Cosine$\uparrow$, 
 S$\phi$rensen $\downarrow$) for each compared algorithm on the controlled  dataset (with b=0.2). For the LDL algorithm $\mathcal{F} \in\{$ AA-BP, AA-KNN, CPNN, LDSVR,PT-Bayes $\}$, the performance of the $\mathcal{F}$-LSag is compared against that of  $\mathcal{F}$, $\mathcal{F}$(G), $\mathcal{F}$(I) represent LDL algorithm training with ground-truth LD and noise LD respectively.  }
\label{recoovertable}
\end{table*}

\begin{table}[!h]\centering
\setlength{\tabcolsep}{1mm}
\begin{tabular}{cccccc}
\hline
\multicolumn{2}{c}{$\mathcal{F}$-LSag} & \multicolumn{4}{c}{Evaluation metric} \\ \cline{3-6} 
\multicolumn{2}{c}{vs} & Chebyshev$\downarrow$ & clark$\downarrow$ & cosine$\uparrow$ & S$\phi$ren$\downarrow$ \\ \hline
\multirow{5}{*}{$\mathcal{F}$} & AA-BP & win[4.88e-04] & win[4.88e-04] & win[4.88e-04] & win[4.88e-04] \\
 & AA-KNN & win[4.88e-04] & win[4.88e-04] & win[6.84e-03] & win[4.88e-04] \\
 & CPNN & win[4.88e-04] & win[4.88e-04] & win[4.88e-04] & win[4.88e-04] \\
 & LDSVR & win[1.46e-03] & win[9.77e-04] & win[9.77e-04] & win[9.77e-04] \\
 & PT-Bayes & win[4.88e-04] & win[4.88e-04] & win[4.88e-04] & win[4.88e-04] \\ \hline
\end{tabular}
\caption{ Wilcoxon signed-rank test between $\mathcal{F}$-LSag and $\mathcal{F}$ in terms of (Chebyshev$\downarrow$, clark$\downarrow$,  Cosine$\uparrow$,  S$\phi$rensen$\downarrow$ ). Significance level $\alpha$=0.05.}
\label{wilcoxon}
\end{table}

\subsection{Further Analyses}

\subsubsection{Parameter sensitivity analysis}: The impact of different hyper-parameters  $\alpha$ and $\beta$ in Eq. (\ref{finnalloss}) on the prediction of experimental results is shown in Figure \ref{canshu1}.  As depicted in Figure 3, the trade-off parameters $\alpha$ and $\beta$, which control the strength of error and preserving the instance correlation topological structure, respectively, do indeed influence the performance of our method. However, the proposed model is quite robust to the those two hyper-parameters, i.e., the Chebyshev distance and S$\phi$rensen distance are relatively stable  as the parameter value changes within a reasonable range, which serves as a desirable property in using the proposed approach. Additionally, the impact of different parameters $\kappa$ and $\nu$ on the prediction results is illustrated in Figure \ref{canshu2}. As shown in Figure \ref{canshu2}, parameters $\kappa$ and $\nu$,  which control the strength of the term between the error of the predicted results and the recovered LD, and the term controlling the alignment between the predicted LD and the ideal LD, respectively, do indeed affect the performance of our method. However, our method still remains stable within a certain range.

\subsubsection{Visualizing Experimental Results }
To better understand our algorithm, we have visualized a portion of the prediction results, as shown in Figure \ref{keshihua}. The first column displays the representative Label Distribution (LD) samples from M2B, RAF-ML, Flickr, and Nature-Scene datasets. The second column  to the last columns present the predicted LDs using different algorithms. In each figure, the x-axis represents various labels, and the y-axis indicates the descriptiveness of the corresponding labels. From Figure \ref{keshihua},  we have the following observations:
\begin{itemize}
    \item When training with Inaccurate Label Distributions (ILDs), algorithms such as AA-BP, CPNN, LSVR, and LDLSF struggle to accurately predict the Label Distribution (LD) for unseen instances. Specifically, they fail to capture the descriptiveness of each individual label and the relative importance ranking among the label descriptiveness.
    \item AA-KNN, LCLR, LDLLC, and PT-Bayes can capture the relative magnitudes between different labels in their predictions; however, the descriptiveness of each individual label is still inaccurate. This is because these algorithms do not consider the noise present in the Label Distribution (LD) while learning the model.
    \item Our algorithm not only accurately predicts the Label Distribution (LD) for each instance but also effectively predicts the ranking of descriptiveness corresponding to different labels. This is because we consider the noise present in the label distribution before learning the classification model.
\end{itemize}

\subsection{Collaboration with ther LDL Algorithms}

In this section, we discuss the scalability of our method, specifically whether the performance of different LDL algorithms can be improved when facing inaccurate label distributions by recovering the ideal LD through the recovery model. The experimental setup is as follows. Prior to training, we use our recovery model (Eq .(\ref{finnalloss})) to recover the ideal LD. Then, we use the recovered LD for training and finally test on the real data. Note that this setup is consistent with the previous settings. The experimental results are presented in Table \ref{recoovertable}. Here, we use 4 datasets to validate whether the recovered LD can help other LDL algorithms improve their performance when faced with ILD. Here, we used two human face datasets, SJAFFE and SBU-3DFE, as well as two yeast datasets, Yeast-alpha and Yeast-cdc. As shown in Table \ref{recoovertable},  $\mathcal{F}$(I) denotes the performance an LDL algorithm trained  on the noise LD, and  $\mathcal{F}$-LSag indicates the performance of that LDL algorithm trained on the recovered LD by our approach. We also show the performance of different LDL algorithms trained on  the ground-truth label distribution (i.e., $\mathcal{F}$(G)), which can be regarded as the performance  upper bound.  To analyze whether there are statistical performance gaps among $\mathcal{F}$(I) and $\mathcal{F}$-LSag, Wilcoxon signed-rank test \cite{jmlr/Demsar06}, which is a widely-accepted statistical test for comparisons of two algorithms over several datasets, is employed. Table \ref{wilcoxon} summarizes the statistical test results and the p-values.   Based on the above results, observations  can be made as follows:\begin{itemize}
\item Noisy LD can cause a significant degradation in the performance for different  LDL algorithms, so it is necessary to address the issue of learning with ILD.

    \item $\mathcal{F}$-LSag is statistically superior to the $\mathcal{F}$ in all cases (4 datasets and four metrics) , suggesting the effectiveness of our approach. This is because accurate supervision information can guide more precise model training.
    \item The performance of $\mathcal{F}$-LSag  is quite close to the  $\mathcal{F}$(I) on different LDL algorithms, indicating our approach can recover high-quality LD from the noisy LD.
\end{itemize}

\section{Conclusion}
This paper investigates the problem of inaccurate label distribution learning  for the first time. To be specific, we treat the noisy LD matrix as the liner combination of an ideal LD matrix and an error label  matrix, and separates them by a novel adaptive graph-regularized  low-rank  and sparse  decomposition model. Then, we use  ADMM to efficiently optimize  the proposed model. The recovered LD are taken into consideration when inducing a LD predictive model for LDL, achieved through the utilization of a specialized objective function.  Extensive experiments demonstrate that our method can effectively address the ILDL problem.  

\bibliographystyle{IEEEtran}
\bibliography{Bibliography}

\end{document}